\documentclass{article}

\usepackage[preprint]{neurips_2023}

\usepackage{amsmath,amsfonts,bm}

\def\eqref#1{equation~\ref{#1}}

\def\1{\bm{1}}

\DeclareMathAlphabet{\mathsfit}{\encodingdefault}{\sfdefault}{m}{sl}
\SetMathAlphabet{\mathsfit}{bold}{\encodingdefault}{\sfdefault}{bx}{n}

\newcommand{\R}{\mathbb{R}}

\usepackage{times}
\usepackage{wrapfig}
\usepackage{caption}
\usepackage{subcaption}
\usepackage{floatrow}
\usepackage{titletoc}
\usepackage{enumitem}
\setlist[itemize]{noitemsep} 

\usepackage{multirow}
\usepackage[utf8]{inputenc} 
\usepackage[T1]{fontenc}    
\usepackage{hyperref}       
\usepackage{url}            
\usepackage{graphicx}
\usepackage{subcaption}
\usepackage{booktabs}       
\usepackage{amsfonts}       
\usepackage{nicefrac}       
\usepackage{microtype}      
\usepackage{xcolor}         
\usepackage{multicol}
\usepackage{formatmacros}

\def\sff{I\mkern-3muI}
\let\norm\abs
\def\sff{\mathord{\mathrm{I\mkern-1muI}}}
\def\projt{{\scriptsize\rotatebox[origin=c]{180}{$\perp$}}}
\def\projn{\perp}

\begin{document}

%
%
%
%
%
%
%
%

\floatsetup[table]{capposition=above}
\title{The geometry of neural network decision boundaries}

%

\author{
    Dylan M Paiton\thanks{Equal contribution. Code:~\url{github.com/bethgelab/adversarial_subspaces}} \\
    \texttt{dpaiton@gmail.com} \\
    \And
    David Luis Schultheiss$^\ast$ \\
    \texttt{david.l.schultheiss@gmail.com} \\
    \And
    Matthias K{\"u}mmerer \\
    \texttt{matthias.kuemmerer@bethgelab.org} \\
    \And
    Zac Cranko \\
    \texttt{zachary.cranko@uni-tuebingen.de} \\
    \And
    Matthias Bethge \\
    \texttt{matthias.bethge@bethgelab.org} \\
    University of T{\"u}bingen \\
}


\maketitle

\begin{abstract}
Artificial neural networks (ANNs) are built using well-understood mathematical operations, yet their complex nature poses significant challenges to intuitively understanding their decision-making processes.
This work introduces a method for visualizing and measuring ANN decision boundary geometry directly in the high-dimensional input space, avoiding distortions found with traditional projection-based techniques.
We analyze boundary curvatures and distances in diverse inputs, architectures, and training methods.
Boundary distances grow smoothly and slowly with consistent curvature as the size of the subspace increases, suggesting that decision boundaries are close to input samples within large, continuous subspaces.
We also quantify how much boundary distances and curvatures reduce near dataset samples after adversarial training.
Training methodology has a significantly stronger impact on boundary distance and curvature than architecture or test-set accuracy. 
These findings are contextualized within prior work, supporting or refuting several previously proposed theories of boundary behavior.
Our analytically and empirically supported methodology lays the foundation for future studies to improve network interpretability and understanding.
\end{abstract}

\section{Introduction}
\begin{figure}[t]
  \centering
  \begin{subfigure}[b]{0.49\textwidth}
      \centering
      \includegraphics[width=\textwidth]{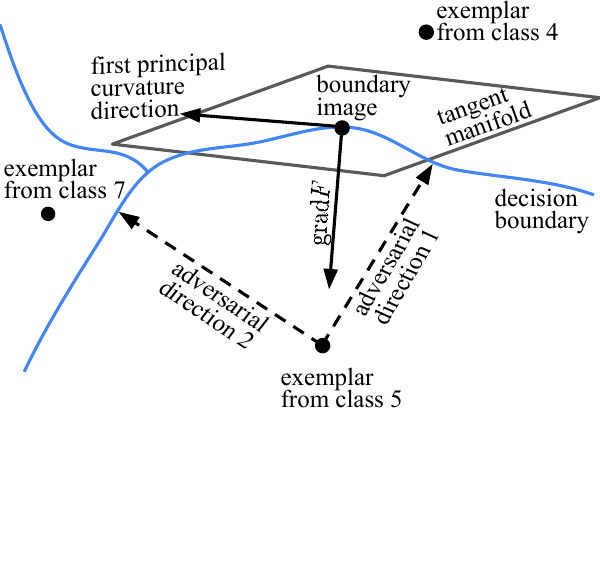}
  \end{subfigure}
  \hfill
  \begin{subfigure}[b]{0.49\textwidth}
      \centering
      \includegraphics[width=\textwidth]{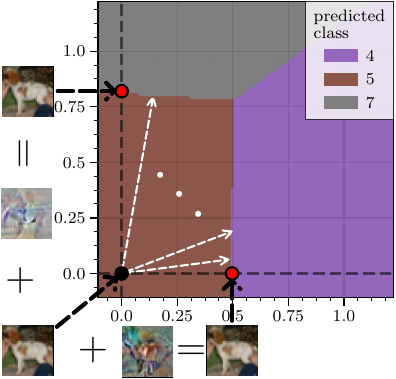}
  \end{subfigure}
  \hfill
  \caption{
  \textbf{The decision boundary for an adversarially trained ResNet-50}.
  (Left) A subspace centered on a dataset image (class 5 exemplar, bottom middle) is defined, with each orthogonal axis corresponding to a direction where the decision boundary is closest.
  Principal curvature is measured at random points within the subspace. 
  (Right) A 2-dimensional subspace is visualized using two adversarial directions, accurately showing the boundary shape and distances to alternate class labels.
  Perturbation images outside the graph are rescaled for visualization, with their $\ell_{2}$ length indicated on the axes.
  Each color represents a class (e.g., deer: class 4, dogs: class 5, horses: class 7).
  White arrows indicate sampling directions within the subspace to locate the boundary.
  }
  \label{fig:diagrams}
  \vspace{-8pt}
\end{figure}

An artificial neural network (ANN) used for classification can be described entirely in terms of its category decision boundaries in the input space.
Understanding these boundaries is crucial for enhancing robustness and interpretability, particularly for applications in high-stakes domains.
However, measuring or visualizing those boundaries is challenging due to the high-dimensional and non-linear relationship between the network's inputs and outputs.
A common approach to understanding these boundaries is to visualize them by projecting the decision surface onto a two-dimensional plane \citep{yu2018interpreting, rodrigues2019constructing}.
Although informative, this suffers from limitations: high-dimensional points can map to the same projected location, the distances between points are distorted, and the projections lack invertibility back to the original high-dimensional decision space \citep{feraud2002methodology, rauber2016visualizing}.
An alternative strategy investigates the decision boundary directly in the high-dimensional input space.
Early work in this direction \citep{golland2001discriminative, baehrens2010how} laid the foundation for the modern focus on adversarial attacks \citep{biggio2013evasion, szegedy2014intriguing}.
Examples produced by adversarial attacks are a worst-case demonstration of a classifier's inability to gracefully cope with identity-preserving shifts or distortions of its input \citep{gilmer2019adversarial}.
To construct one, an adversary perturbs the ANN input in a small but specific way such that the output changes significantly.
The perturbation is optimized to be as close as possible to the ANN's decision boundary.

Adversarial examples provide context for the ultimate goal of improving interpretability and our understanding of classifier behavior by measuring and visualizing the output space near unaltered dataset samples.
Previous work has explored ANN responses in the proximity of adversarial examples \citep{warde2016adversarial, tabacof2016exploring, he2018decision, tramer2017space, swirszcz2019visualizations, karimi2022decision}, or within hand-picked two-dimensional input domains \citep{humayun2023splinecam}.
However, these methods analyze one or two boundary dimensions at a time, leaving key questions unanswered about the high-dimensional structure of the boundary.
For example:
How many orthogonal directions from an input sample maintain close proximity to the decision boundary?
How does the boundary behave between these directions?
Does the boundary distance increase smoothly or does it fluctuate with increased search space dimensionality?
What is the high-dimensional distribution of class label regions around dataset inputs?

In this work, we introduce a novel approach that quantitatively analyzes the geometry of ANN decision boundaries in high-dimensional subspaces centered on dataset samples. Our method provides the first evidence that decision boundaries are close to natural inputs within large subspaces across architectures and datasets, supporting the hypothesis that adversarial examples occupy dense regions of input space \citep{goodfellow2014explaining, tabacof2016exploring, moosavi2017universal, he2018decision}. Specifically, we propose an untargeted adversarial attack framework coupled with a principled method to measure high-dimensional boundary curvature. Our approach corrects inaccuracies in prior boundary curvature estimates and enables more precise 2D visualizations.

Our findings reveal a large subspace around input samples where the decision boundary remains only marginally farther away than at its closest point. Moreover, we show that boundary distances increase smoothly as the dimensionality of the subspace grows. We further demonstrate that adversarial training, widely considered the state-of-the-art defense against adversarial attacks \citep{gowal2020uncovering, bai2021recent}, not only increases boundary distance but also reduces boundary curvature, supporting prior hypotheses \citep{moosavi2018robustness}. Additionally, we provide new evidence that adversarial training affects class label composition near natural inputs, indicating reduced semantic information in label proximity.

In summary, we contribute the first quantitative analysis of the geometry of nonlinear ANN decision boundaries in high-dimensional subspaces centered on dataset samples, a novel curvature measurement method that corrects errors in prior approaches, and new insights into adversarial training’s impact on boundary smoothness and label distributions.
Our experiments confirm consistent trends across datasets, architectures, and generalization performance, offering a principled foundation for understanding decision boundaries and improving interpretability in deep learning.

\begin{figure}[t!]
  \centering
  \includegraphics[width=\textwidth]{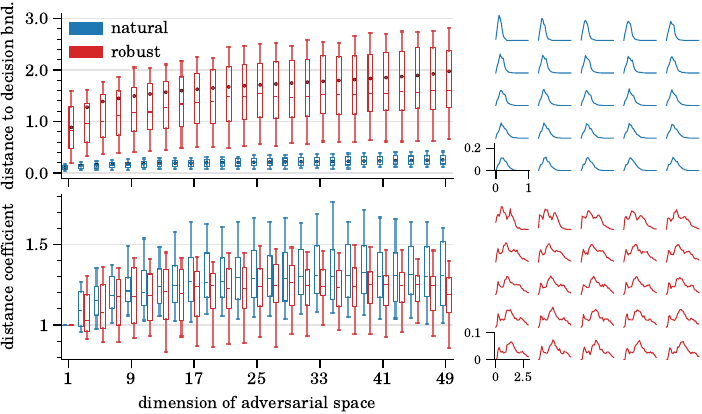}
  \caption{\textbf{Boundary distances.} 
  The distance from a test set image to the decision boundary grows sub-linearly with increasing subspace dimensionality and remains consistent across model architectures. 
  (Left, top) Box plots show decision boundary distances for uniformly sampled directions within n-dimensional subspaces (horizontal axis), with dots indicating distances along the subspace axes, averaged across images.
  Boxes show interquartile ranges, whiskers represent the 10\% and 90\% percentiles, and dashed lines show random sample means.
  Every other dimension up to 50 is shown.
  (Left, bottom) Similar to the top panel, but the vertical axis shows the largest boundary distance within the subspace divided by the largest boundary distance along the subspace axes.
  Low values indicate that the boundary remains close to the input sample between the axes. 
  (Right) Histograms of boundary distances for randomly sampled directions within subspaces, with dimensionality increasing across subplots (odd-numbered dimensions are shown).
  The mean shifts rightward and variance increases proportionately with dimensionality, suggesting a smooth growth in distance with increasing dimensionality.
  Line colors represent training schemes as per the legend.
  }
  \label{fig:cifar_perturbation_distances}
  \vspace{-8pt}
\end{figure}

\section{Methods}
We aim to describe both the proximity and the curvature of the decision boundary within high-dimensional subspaces centered on test-set inputs.
To demonstrate the robustness of our findings, we compare two ResNet architectures \citep{he2016deep, gowal2020uncovering} trained on the CIFAR-10 dataset \citep{krizhevsky2009learning}, which differ in their training parameters and generalization performance.
We then perform a more detailed curvature analysis on the smaller of the two architectures.
Additional dataset and model details are provided in Appendix \ref{app:datasets_models}.
Control experiments that validate consistency across datasets and random seeds are provided in Appendices \ref{app:add_cifar}, \ref{app:mnist}, and \ref{app:seeds}.
Our experiments were conducted on 100 test images, chosen to ensure equal label representation (10 images per label), all of which were correctly classified by the networks under study.

\subsection{Defining subspaces with minimum boundary distance} \label{sec:decomposition}

Consider a neural network $f$ that receives an $n$-pixel input, $x \in \mathbf{R}^{n}$, and produces an output vector of normalized scores associated with $l$ label categories, $\hat{y} \in \mathbf{R}^{l}$.
We define a subspace centered on the input using a set of orthogonal vectors $\Delta = [\delta_{1}, \delta_{2}, \ldots, \delta_{m}], m \le n$.
The method for finding this set must jointly enforce a bounded input, $x + \delta_{m} \in [0,1]^n$, orthogonality, $\langle \delta_{k} , \delta_{m} \rangle = 0,\ \ \forall k<m$, and misclassification
\footnote{
    Since we are not targeting any specific class, this formulation is needed to avoid a completely uniform output distribution (i.e. no class wins).
}
, $\operatornamewithlimits{max}\limits_{j\ne i} f_{j}(x + \delta) - f_{i}(x + \delta) > 0$, where $i$ is the correct class index.
Additionally, the perturbation vector $\delta_{m}$ should have minimal length,  $||\delta_{m}||_{p}$, where $||\delta||_{p} = (\sum_{i}|\delta|^{p})^{\frac{1}{p}}$ is the $l_{p}$ norm.
While solving this exactly for deep networks is infeasible, it can be approximated using an untargeted adversarial attack objective.
We define $\mathbf{\ell} \defeq \frac{1}{L_{CE}(x + \delta) } + \kappa \cdot ||\delta||_{p}$, which includes the cross-entropy loss $L_{CE}(x) \defeq - \log{\frac{\text{exp}f_{i}(x)}{\text{exp}\sum_{j}f_{j}(x)}}$.
Here, $\kappa$ is a penalty coefficient optimized via binary search at run time.
The components combine to give the full objective function as implemented:

\begin{equation}\label{eq:opt_objective}
\begin{aligned}
    &\operatorname{minimize}    && \frac{1}{L_{CE}(x + \delta_{m})} + \kappa \cdot ||\delta_{m}||_{p}\\
    &\text{subject to}           && \operatornamewithlimits{max}\limits_{j\ne i} f_{j}(x + \delta_{m}) - f_{i}(x + \delta_{m}) > 0\\
    &                           && x+\delta_{m} \in [0,1]^n\\
    &                           && \langle \delta_k , \delta_m \rangle = 0,\ \ \forall k<m.
\end{aligned}
\end{equation}

which we solved using the interior-point optimizer implemented by Ipopt \citep{wachter2006implementation}.
This is in line with prior work that uses the cross-entropy loss in an adversarial objective \citep{szegedy2014intriguing, carlini2017towards}, but modified to match an untargeted attack objective \citep{tabacof2016exploring, finlay2019logbarrier}
\footnote{
    Note that this is different from ``untargeted'' adversarial attacks that target the second-most-likely output class \citep{he2018decision, chen2018ead}.
}
.

After iteratively minimizing this objective, the first vector points to the closest decision boundary, while subsequent vectors correspond to orthogonal directions with increasing distances to the boundary.
The number of vectors is limited by computational constraints, but the method can theoretically scale up to the input dimensionality.
This greedy approach provides a subspace that is constrained by an empirical upper bound for boundary distances.

We used sampling to find boundary distances at locations between subspace axes.
Specifically, within each subspace, we sampled 100 random directions using a linear combination of normally distributed variables, a method known to provide even sampling in high-dimensional spaces \citep{muller1959note}.
We then used a binary search to find the decision boundary along each sampled direction.
All distances are measured using the $\ell_{2}$-norm.
Improved methods for finding orthogonal adversarial perturbations, sampling, or searching for the boundary could further refine our measurements.
However, we do not expect such an improvement to qualitatively modify our conclusions.

\begin{figure}[t!]
  \centering
  \includegraphics[width=\textwidth]{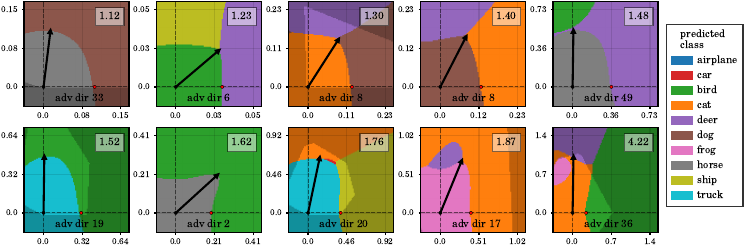}
  \caption{\textbf{Boundary visualizations.} 
  Each subplot shows a two-dimensional cross-section of the decision space for a naturally trained network, centered on a random test set image. 
  The cross-section corresponds to the subspace with the largest distance coefficient (top-right corner), calculated as the ratio of the largest boundary distance within the subspace (black arrow) to the largest adversarial perturbation defining the subspace (horizontal axis, with dimension number indicated). 
  Red dots mark the described adversarial perturbations.
  Shaded areas represent regions outside the image bounds. 
  Unlike projections, this visualization preserves high-dimensional spatial relationships, ensuring accurate representation of boundary geometry.
  }
  \label{fig:boundary_visualizations}
  \vspace{-8pt}
\end{figure}

\subsection{Measuring decision boundary curvature}\label{sec:curvature_short}
Neural network decision boundaries can be sufficiently described as high-dimensional Riemannian manifolds.
Building on this perspective, we derive a formula for quantifying the curvature of a decision boundary both in its native dimensionality as well as within any subspace containing the boundary.

Let $f_i, f_j:\R^m\to\R$ be the functions corresponding to the $i$-th and $j$-th logits of a neural network,~$f$.
Define $F_{ij} \defeq f_i - f_j$.
Then the \emph{$(i,j)$-decision boundary} is

\begin{gather}\label{eq:boundary_def}
    B_{ij}\defeq \setcond{ x\in \R^n} { F_{ij}(x) = 0 }  = \setcond{ x\in \R^n} { f_i(x) = f_j(x)}.
\end{gather}\noeqref{eq:boundary_def}

The set $B_{ij}$ is independent of the sign of $F_{ij}$ (i.e., $B_{ij} = B_{ji}$), though the ordering will affect the sign of curvature.
We follow the convention established earlier and in \citet{moosavi2018robustness} by associating $i$ with the correct class label for a point $x$ and $j$ with an incorrect label.
This ensures that positive curvature of $B_{ij}$ near $x$ corresponds to the boundary bowing inward towards $x$, while negative curvature indicates the boundary bowing outward (see Figure \ref{fig:diagrams} for an example of positive curvature).
If the curvature is zero near $x$, then $B_{ij}$ is locally affine, or flat.

We apply principal curvature decomposition at 50 boundary points to quantify the extent of linearity.
This decomposition, performed on a level-set of the decision space (generally of codimension 1), yields $n-1$ scalars and vectors called the \emph{principal curvatures} and \emph{principal curvature directions}, respectively.
Conceptually analogous to principal component analysis, which organizes the space as an orthogonal basis of directions of maximal variance, principal curvature analysis organizes the (tangential) space of the decision boundary into orthogonal directions of maximum curvature.
Specifically, the principal curvatures and associated directions correspond to the eigenvectors and eigenvalues of a linear operator $s:\R^m\to\R^m$ called the \emph{shape operator}.

To investigate curvature within a subspace, we define the \emph{subspace pullback} of the shape operator:

\begin{gather}\label{eq:subspace_pullback}
    \rbr{P^\trsp}^*s \defeq P s P^\trsp,
\end{gather}\noeqref{eq:subspace_pullback}

where $P\in\R^{m \times n}$ is an orthogonal projection matrix, and the image of $P^\trsp$ spans the subspace of $\R^m$.
This formulation restricts the shape operator to a linear subspace of the tangential space.
It also allows for a direct comparison of curvature between a subspace and the full response space.
The mathematical motivation for this definition is rooted in differential geometry, with a full treatment provided in \autoref{app:curvature_long}.

\begin{figure}[t!]
  \centering
  \includegraphics[width=\textwidth]{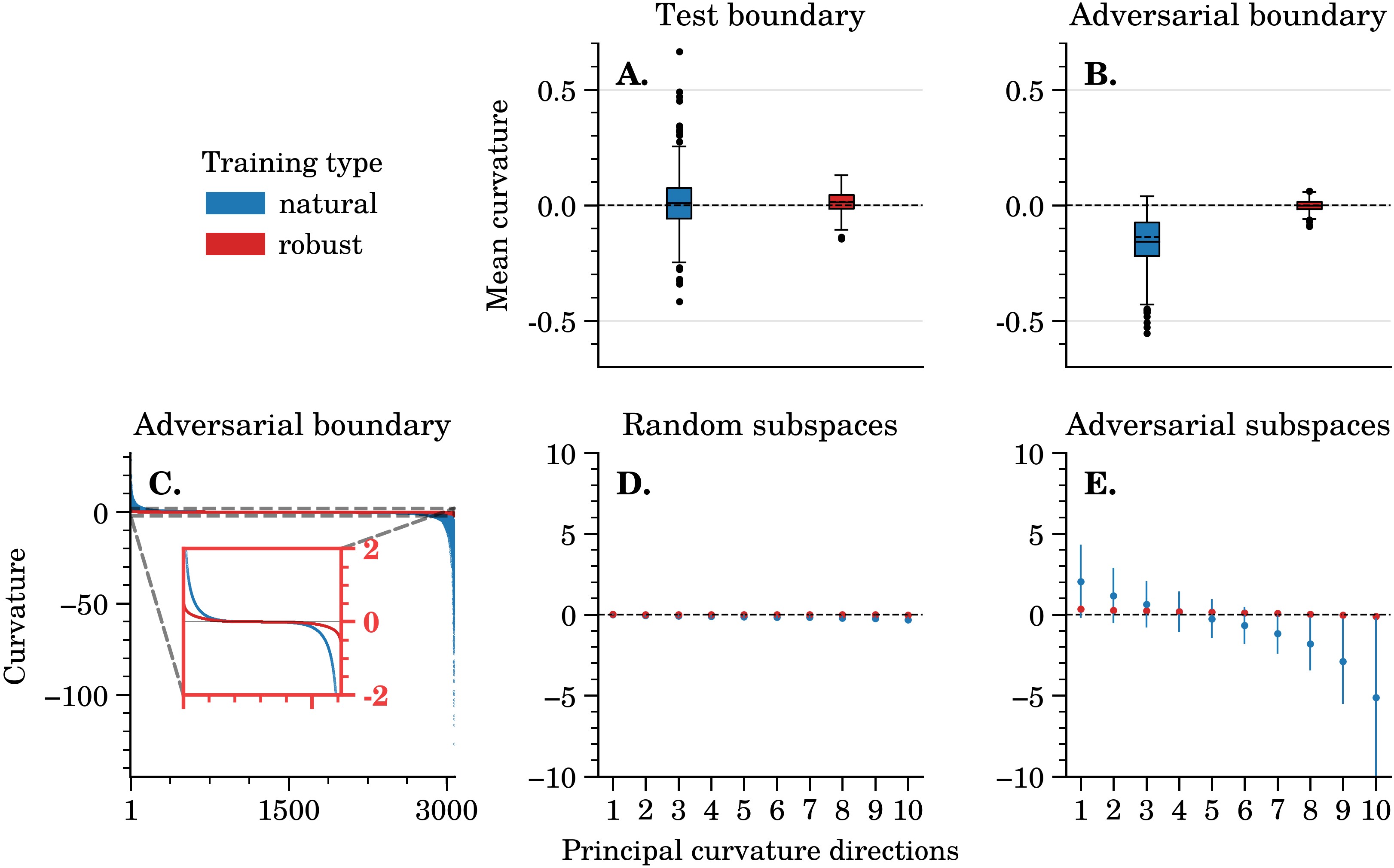}
  
  \caption{\textbf{Boundary curvature.} 
  Curvature data for 50 correctly classified test images, with evenly distributed labels for the first 10 dimension numbers.
  (A, B) Mean curvatures per boundary point are smaller for boundaries separating two test images (``test boundary'') compared to those separating a test image and an adversarial image (``adversarial boundary''). 
  (C) Scatter plot of curvature profiles for all images and directions under the adversarial boundary condition.
  The red inset shows the full dimensionality (same horizontal axis as the main plot) with a zoomed vertical axis, highlighting that adversarially trained boundaries are flatter (more linear).
  The data includes all subspace boundary points per test image.
  (D, E) Curvature of adversarial boundaries within random subspaces (D) and adversarial subspaces (E).
  Adversarial subspaces exhibit more extreme curvature.
  Points show the mean across test images; error bars represent standard deviation.
  }
  \label{fig:curvature_analysis}
  \vspace{-8pt}
\end{figure}

\section{Results}\label{sec:results}
The distance to the decision boundary increases with each additional dimension of our defined subspace.
The term \emph{dimension number} refers to the ordered axes for the subspace, where the lowest dimension number corresponds to the first axis with the closest boundary point, and higher dimension numbers indicate increasingly distant boundary points.
The perturbations become visually noticeable at approximately dimension number 20 for the robust network and beyond dimension number 50 for the naturally trained network (see Appendix \ref{app:add_cifar} Figure \ref{fig:cifar_examples} for examples).

\subsection{Boundary distances}
The circles in the top left panel of Figure \ref{fig:cifar_perturbation_distances} represent the average boundary distances along each dimension number, computed across test images by greedily minimizing Equation \ref{eq:opt_objective}.
The box plots in the top-left panel of Figure \ref{fig:cifar_perturbation_distances} display the distribution of decision boundary distances across sampled directions and test images.
These results confirm that the mean initial perturbation distances align with those reported in prior work on adversarial attacks \citep{madry2018towards} and decision boundary distance estimations \citep{ding2020mma} (also see Appendix Table \ref{tab:cifar_accs}).
The bottom-left panel of Figure \ref{fig:cifar_perturbation_distances} shows the distribution of \emph{distance coefficients}, defined as the ratio of the maximum distance to the decision boundary within a subspace to the largest boundary distance along the defining subspace axes (which will be along the axis with the highest dimension number).
For example, a value of 2 indicates that the farthest sampled boundary point is twice as distant as the farthest boundary point along an axis.
Histograms in the right-hand panels of Figure \ref{fig:cifar_perturbation_distances} count sampled boundary distances, with subspace dimensionality increasing from left to right and top to bottom.
They indicate that the decision boundary distances increase smoothly across dimensions, with minimal variation between subspace axes.
Distances were only reported for boundary points that remained within the $\ell_{\inf}$ pixel bounds, covering at least 40\% of the sampled directions for the naturally trained network and 60\% for the adversarially trained network (see Appendix \ref{app:add_cifar} Figure \ref{fig:cifar_out_bnd}).

The distance to the decision boundary grows sub-linearly as the dimensionality of the subspace increases, with the most significant growth occurring in the first few dimensions.
The adversarially trained network consistently exhibits smaller distance coefficients, suggesting less variability in boundary distances within the subspace.

\subsection{Boundary visualizations}
Figure \ref{fig:boundary_visualizations} presents two-dimensional cross-sections of the decision surface.
The horizontal axis in each cross-section corresponds to a specific subspace axis, with the associated dimension number labeled.
The black arrow marks the farthest decision boundary point sampled within the subspace.
The vertical axis is determined through one step of the Gram-Schmidt process, ensuring that the visualized space is spanned by the horizontal axis and the vector pointing to the farthest sampled boundary point.

The cross-sections are ordered by their distance coefficient, displayed in the upper-right corner of each plot.
These visualizations include instances where the decision boundary distance increases smoothly between axes, as well as less frequent cases where the boundary extends significantly farther from the origin between the subspace axes (a pattern also observed in Figure \ref{fig:cifar_perturbation_distances}).

\subsection{Boundary curvature}
Figure \ref{fig:curvature_analysis} presents boundary curvature measurements under various experimental conditions, including different training methods, boundary locations, and subspaces.
Panels A and B display the mean curvature (i.e. the mean of the eigenvalues of the shape operator) for points on the decision boundary.
Boundary points separating unperturbed test set examples from adversarial points (adversarial boundary; Panel B) exhibit higher mean curvature compared to boundary points separating the examples themselves (test boundary; Panel A).
Panel C shows a scatter plot of the principal curvatures (i.e. eigenvalues of the shape operator) at the found boundary points in the high-dimensional input space.
Both positive and negative curvature values are observed at all boundary points.
Panels D and E report principal curvatures within subspaces defined by random vectors and by adversarial vectors, respectively.
Curvature within the found subspaces (Panel E) is notably larger than in random subspaces (Panel D).
For naturally trained networks, the boundary points in both the full input space (Panel B) and the found subspace (Panel E) tend to exhibit more negative mean curvature.
The subspace curvature is measured using the pullback defined in Equation \ref{eq:subspace_pullback}, which establishes bounds on curvature amounts between each principal vector.
Boundary points closest to the clean input can exhibit positive curvature only if the radius of curvature exceeds the distance to the boundary.

We observe consistent curvature measurements when varying model architectures, test-set accuracy, and random seeds for two different datasets (see Appendices \ref{app:mnist} and \ref{app:seeds})
However, differences in training methodology lead to notable variations in boundary geometry for otherwise identical architectures.
Adversarial training linearizes the decision boundary, both between correctly labeled test images and near adversarial examples.
The difference in curvature values between naturally and adversarially trained networks is less pronounced in random subspaces than subspaces found via an adversarial objective.

\subsection{Decision class composition}
\begin{figure}[t!]
  \centering
  \includegraphics{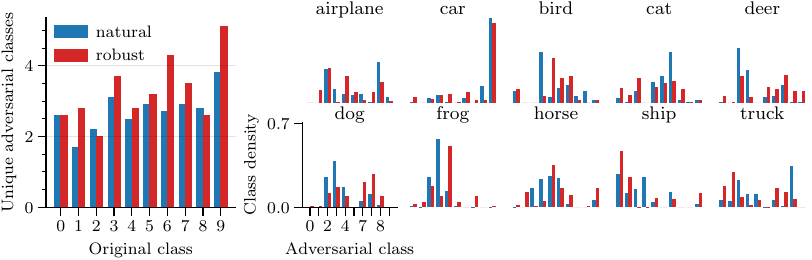}
  \caption{\textbf{Adversarial Class Composition.}
  (Left) Within the subspaces of randomly chosen test images we quantified how many unique adversarial classes exist.
  In most cases adversarial training increases the diversity of alternate classes close to any test image.
  (Right) Each plot shows the distribution of nearby alternate classes for each origin class with natural and adversarial training.
  Adversarial training results in a more even distribution of nearby classes for all categories except bird and ship.
  }
  \label{fig:adv_class_composition}
  \vspace{-8pt}
\end{figure}

Figure \ref{fig:adv_class_composition} presents measurements of the diversity of adversarial classes near test samples.
The leftmost subplot shows the average number of unique adversarial classes found within the first 50 dimensions of the subspace, computed across correctly classified test images.
The remaining subplots display the distribution of adversarial classes within the subspaces, categorized by the original label of the test samples.
Table \ref{table:class_entropy} provides the discrete entropy of each distribution, quantifying the diversity of adversarial classes present.
Adversarially trained networks have higher entropy values and a broader distribution of adversarial classes compared to naturally trained networks.
This suggests that adversarial training leads to an increase in the number of unique adversarial classes near test samples.

\begin{wraptable}{r}{0.4\textwidth}
   \centering
   \vspace{-10pt}
   \begin{tabular}{c c c}
   \\
   \toprule
   Label & Natural & Robust \\
   \midrule
   airplane & 1.70 & \bf{1.81} \\
   car & 1.01 & \bf{1.26} \\
   bird & \bf{1.72} & 1.59 \\
   cat & 1.56 & \bf{1.97} \\
   deer & 1.41 & \bf{1.98} \\
   dog & 1.51 & \bf{1.82} \\
   frog & 1.12 & \bf{1.51} \\
   horse & 1.63 & \bf{1.78} \\
   ship & \bf{1.74} & 1.39 \\
   truck & 1.79 & \bf{1.88} \\
   \bottomrule
   \end{tabular}
   \caption{Entropy of unique proximal adversarial labels per class, averaged across a balanced sampling of correctly labeled test images.}
   \vspace{-10pt}
   \label{table:class_entropy}
\end{wraptable}

\section{Discussion and related work}
\textbf{Adversarial subspaces:}
Previous research has extensively explored subspaces near adversarial perturbations \citep[e.g.,][]{fawzi2016robustness, tabacof2016exploring, liu2017delving, tramer2017space, he2018decision, ma2018characterizing}.
\citet{fawzi2016robustness} studied the nearest error point in random subspaces, while others analyzed classifier behavior through random perturbations near adversarial examples \citep{warde2016adversarial, he2018decision}.
\citet{gilmer2019adversarial} and \citet{hendrycks2019benchmarking} examined worst-case and general out-of-distribution (OOD) errors, highlighting adversarial inputs as examples of the former.

While these works suggest that local neighborhoods around adversarial examples are often adversarial, indicating a high-rank error space, they leave unanswered questions about the multi-dimensional decision boundary near natural inputs.
Our method addresses this gap by identifying subspaces where proximal decision boundary geometry can be measured, leading to key insights:
\begin{itemize}
    \item \textbf{Proximity assessment}: The decision boundary is close to natural inputs within a large subspace, with distances increasing smoothly and slowly as dimensionality grows.
    This contradicts hypotheses that adversarial examples are isolated \citep{szegedy2014intriguing} or caused by sharp boundary protrusions \citep{yu2019new}.
    \item \textbf{Boundary curvature:} 
    Adversarial training linearizes boundaries proximal to training images more than distal boundaries, while naturally trained network boundaries tend to exhibit negative curvature when proximal to natural inputs.
    \item \textbf{Adversarial class composition:} Adversarial training produces a more uniform class distribution, indicating greater utilization of high-dimensional space but less alignment with semantic proximity.
\end{itemize}

Our findings align with studies on adversarial robustness and input dimensionality \citep{amsaleg2017vulnerability, simon2019first}, showing consistent trends across datasets.
They also support evidence for large adversarial subspaces, as suggested by works on combining universal adversarial directions \citep{liu2024data, patracone2024linear}.
Prior work has used decision boundary measurements to interpret neural network behavior \citep{mickisch2020understanding} but did not address multidimensional boundary geometry, which we explore in detail.
Further discussion of prior approaches to defining adversarial subspaces can be found in Appendix~\ref{app:orthogonal_clipping}.

\textbf{Neuron response curvature:}
Analysis of response curvature has a rich history in neuroscience \citep[see][for a review]{gollisch2012iso} and computational models of neural activity \citep[e.g.][]{zetzsche2001nonlinear, rust2005spatiotemporal, golden2016conjectures, cohen2020separability}.
In the context of ANNs, \citet{golden2019selective} employed principal curvature decomposition to study neuron selectivity and invariance, while \citet{paiton2020selectivity} connected neuron response curvature to adversarial robustness.
Our work differs by focusing on decision boundary curvature rather than individual neuron response curvature.

To the best of our knowledge, our work constitutes the first comprehensive measurement of the decision boundary curvature for nonlinear neural networks.
We improve on previous work \citep[][]{fawzi2016robustness, poole2016exponential, moosavi2018robustness, moosavi2019robustness, golden2019selective, mickisch2020understanding, lei2023understanding, sekmen2024manifold} by providing complete details of the derived formulas in \autoref{sec:curvature_short} and demonstrating that our solution is uniquely correct by measuring the level-set curvature of simple Riemannian manifolds in \autoref{app:moosavi_comparison}.
We also corroborate and extend the findings of \citet{moosavi2018robustness}, demonstrating a positive correlation between adversarial vulnerability and decision boundary curvature, with additional comparisons across adversarial subspaces, random subspaces, and the full input space.

\section{Conclusion}\label{sec:conclusion}

We introduce a novel method for measuring the geometry of neural network decision boundaries, presenting the first evidence that decision boundaries are close to natural inputs within large subspaces across architectures and datasets (Figure \ref{fig:cifar_perturbation_distances}).
This supports the hypothesis that adversarial examples occupy dense regions of input space \citep{goodfellow2014explaining, tabacof2016exploring, moosavi2017universal, he2018decision} while refuting alternative explanations \citep{szegedy2014intriguing, yu2019new}.
Our method improves the accuracy of 2D visualizations (Figure \ref{fig:boundary_visualizations}), defining subspaces (Appendix~\ref{app:orthogonal_clipping}), and boundary curvature estimation (Appendix \ref{app:moosavi_comparison}).
We provide new evidence that adversarial training linearizes decision boundaries (Figure \ref{fig:curvature_analysis}) \citep{li2019preventing, moosavi2019robustness, terjek2020adversarial, sakar2020enforcing, zhang2019theoretically, wu2020adversarial}, along with information on its impact on class label composition near natural inputs (Figure \ref{fig:adv_class_composition}). 

Applications of our work include semi-supervised relabeling \citep{benato2018semi}, architecture design \citep{rauber2018projections}, and robustness enhancement \citep{tian2021detecting, sekmen2024manifold}.
Furthermore, our framework supports quantitative, falsifiable interpretability, essential for deploying neural networks in high-stakes applications \citep{ribeiro2016why, vlassopoulos2020explaining, doshi2017towards, leavitt2020towards}.
Future directions include large-scale analysis across architectures, particularly with non-smooth activations \citep[e.g., ReLU or stochastic, ][]{li2019certified, dapello2020simulating}, and investigating relationships between adversarial directions, boundary curvature, overfitting \citep{tanay2016boundary}, and adversarial feature perceptual quality \citep{ilyas2019adversarial}.

\newpage
\section*{Author Contributions}
All figures and the first draft were written by D.P. and D.S., although all authors contributed to revisions.
The subspace finding algorithm was developed by D.S., and the optimization procedure was developed by D.S. and M.K.
All curvature related experiments were conceived by D.P., and written by D.P. and M.K.
All other experiments were written and conducted by D.S.
The theoretical analysis of boundary curvature was conducted by Z.C., D.P., and M.K.
The idea to investigate subspaces where the decision boundary is closest was first proposed by M.B.
The project was led by M.B.

\section*{Acknowledgments}
We would like to thank James Golden for early guidance, Sven Gowal for providing extra trained networks, Wieland Brendel and Yash Sharma for valuable feedback, as well as Jonas Rauber for his contributions on early iterations of the project. 
Our work was funded by the Deutsche Forschungsgemeinschaft (DFG, German Research Foundation) under Germany’s Excellence Strategy – EXC number 2064/1 – Project number 390727645; and The German Federal Ministry of Education and Research (BMBF): Tübingen AI Center, FKZ: 01IS18039B.

\bibliography{main}

\newpage
\appendix

\startcontents[sections]
\printcontents[sections]{l}{1}{\setcounter{tocdepth}{1}}

\section{Full derivation for hypersurface curvature}\label{app:curvature_long}
We use definitions that are commonly found, and our notation follows \citet{leeIntroductionSmoothManifolds2013,leeIntroductionRiemannianManifolds2018} for differential geometry, and \citet{magnusMatrixDifferentialCalculus1999} for linear algebra.

\subsubsection{Definition of terms}
In order to derive our main curvature expressions for a graph submanifold, we first introduce some notation and recall some concepts.

\paragraph{Tangent Spaces}
If $M$ is an $n$-dimensional smooth manifold then the \emph{tangent space at $p\in M$} is a vector space $\tangb_pM$ of derivations at $p$, which is isometric to $\R^n$.
In particular, $\tangb_pM$ is the set of directional derivative operators acting on the smooth real functions $f:M\to \R$. 

The disjoint union of all of the tangent spaces is the \emph{tangent bundle} $\tangb M \defeq \setcond{ \set{p}\times \tangb_p M}{p\in M}$.
If we equip $M$ with a smooth, bilinear, positive definite, symmetric form $g:\tangb M\times\tangb M \to \R$ then $(M, g)$ is said to be a \emph{Riemannian manifold} and $g$ the \emph{Riemannian metric tensor}.

When it is unambiguous, denote by $\secsm(E)$ the smooth mappings $M\to E$ (for some topological space $E$).
Most often we will refer to $\secs(M) \defeq \secsm(\tangb M)$, and represent elements of $\secs(M)$ with the letters $X,Y,Z$. 

\paragraph{Differentials}

If $M$ and $N$ are manifolds and $F:M\to N$ is differentiable, then the \emph{differential of $F$ at $p\in M$} is the linear mapping denoted $\dv F_p: \tangb_p M \to \tangb_p N$,
where the action on tangent vectors is defined by 
\begin{gather}
  \forall{p\in M}\forall{v\in\tangb_pM}\forall{f\in\contf^{1}(N)}\dv F_p(v)f = v (f\circ F)(p).
\end{gather}
That is, the rate of change of the function $f\circ F$ in the direction $v\in\tangb_pM$ at the point $p$.
As is common practice, we often omit the $p$ subscripts for clarity.
The linear operator $\dv f$ is a linear functional when $N$ in the above definition is $\R$, and at each $p\in M$ there exists a vector $\grad f_p \in \tangb_p$.
So that
\begin{gather}
  \forall{p\in M}\forall{v\in\tangb_p M} \dv f_p(v) = g(\grad f_p, v),
\end{gather}
and $\grad f\in\secs(M)$ is the associated vector field.

\paragraph{Connections}
A \emph{connection} is an operator $\connar:\secs(M)\times \secs(M) \to \secs(M)$, linear in its first argument, and satisfying a product rule in its second.
This defines a connection in each tensor bundle,\footnote{Including the $(0,0)$-tensor bundle, that is, the smooth real functions on $M$.} which we overload with $\connar$ \citep[Prop.~4.15]{leeIntroductionRiemannianManifolds2018}. Importantly for us, we have $\connar_X f = Xf$ for all $X\in\secs(M)$ (for any connection).\footnote{That is, the action of a vector field through the connection agrees with the ordinary directional derivative.}
A Riemannian manifold always possesses a \emph{Levi--Cevita connection} \citep[\S5]{leeIntroductionRiemannianManifolds2018}, denoted by $\conn$, which is the unique connection satisfying
\begin{align}
  \forall{X,Y,Z\in\secs(M)}\conn_Xg\rbr{Y,Z} = g\rbr{\conn_XY, Z} + g\rbr{Y, \conn_XZ}.
\end{align}

To $f\in\contf^2(M)$ we associate the Hessian operator $\conn^2f$, which is a symmetric $(0,2)$-tensor field \citep[Ex.~4.22]{leeIntroductionRiemannianManifolds2018} with $\conn^2f(X,Y)
\defeq \conn_X \dv f(Y)$ for $X,Y\in\secs(M)$. 
We have \citep[via][Prop.~4.21]{leeIntroductionRiemannianManifolds2018}
\begin{align}
    \forall{X,Y\in\secs(M)}\conn_X \dv f(Y) = \conn_X\rbr{\conn_Y f} -  \conn_{(\conn_XY)}f
    = X (Yf) - (\conn_X Y)f.\label{eq:hess_expr}
\end{align}
Using the compatibility of $\conn$ with the metric we arrive at the following expression for the Hessian, which will be useful later
\begin{align}
    \nabla^2f(X,Y)
    &\overseteqref{eq:hess_expr}{=} X (Yf) - (\conn_X Y)f
    \\&= \underbrace{X g\rbr{\grad f, Y}}_{\conn_X g\rbr{\grad f, Y}} 
        - g\rbr{\grad f, \conn_X Y}
    \\&= g\rbr{\conn_X \grad f, Y} + g\rbr{\grad f, \conn_X Y} - g\rbr{\grad f, \conn_X Y}
    \\&= g\rbr{\conn_X \grad f, Y}. \label{eq:hessian}
\end{align}

\subsubsection{Submanifolds}

If $(M,g)$, $(\tilde M, \tilde g)$ are two Riemanian manifolds with $M\subseteq\tilde M$, then an injection $\iota:M\hookrightarrow \tilde M$ is called an \emph{inclusion mapping}. If the inclusion is a smooth isometry, that is
\begin{gather}
  \forall{u,v\in\tangb M} g(u,v) = \tilde g(\dv\iota(u), \dv\iota(v)),
\end{gather}
equivalently, $g = \iota^*\tilde g$, then $M$ is called a \emph{Riemanian submanifold of $\tilde M$}.

Using the inclusion map differential, we embed $\tangb M$ in $\tangb \tilde M$.
If $M$, $\tilde M$ are of dimensions $m<n$, then $\tangb_p M$ occupies an $m$ dimensional subspace of $\tangb_p \tilde M$ for each $p\in M$.
The \emph{normal bundle} is formed from the orthogonal compliment of $\tangb_p M$ at each $p \in M$, and is defined in a similar way to the tangent bundle, denoted $\normb M$ and having a dimension of $n -m$.
For vectors $v\in\tangb \tilde M$ we have the decomposition $v = v^\projt + v^\projn$ with $v^\projt \in \tangb M$ and $v^\projn \in \normb M$.

Some care must be taken when taking covariant derivatives in $\tangb M$ with the ambient connection, formally 
\begin{align}
    \forall{X,Y\in\tangb M}\tilde\conn_X Y \defeq \tilde\conn_{\overline{\dv\iota (X)}} \overline{\dv\iota (Y)},
\end{align}
where the overlines denote arbitrary smooth extensions of $\dv\iota (X)$ and $\dv\iota (Y)$ to open neighbourhoods of $M$ in $\tilde M$.

\paragraph{Second Fundamental Form} Let $M\subseteq \tilde M$ be an embedded sumbanifold. The \emph{second fundamental form} is the (bilinear) mapping $\sff: \secs(M)\times\secs(M)\to\secsm\rbr{\normb M}$ with $\sff(X,Y) \defeq \rbr1{\tilde\conn_XY}^\perp$.
If we equip $M$ with a normal field $N\in \secsm(\normb M)$, then the \emph{scalar second fundamental form} is
\begin{gather}
    \sff_N(X,Y) \defeq \tilde g\rbr{N, \sff\rbr{X,Y}}.
\end{gather}
For the remainder we assume $N$ is fixed. The \emph{shape operator} $s:\secs(M)\to\secs(M)$ is the endomorphism obtained from $\sff_N$ that satisfifes
\begin{gather}
  \forall{X,Y\in\secs(M)} \sff_N(X,Y) = g(sX, Y),\label{eq:shape_operator_def}
\end{gather}
that is, by raising an index.

\paragraph{Curvature} Let $s$ be the shape operator associated to a normal field on an embedded submanifold $M$ of dimension $n$.
Then at $p\in M$, the (real) eigenvalues of $s$ are called \emph{principal curvatures} and the eigenvectors are called \emph{principal directions} at $p$.
The \emph{Gaussian curvature} is the product of the principal curvatures, $\det(s)$, and the \emph{mean curvature} is their average, $\frac1n\trace(s)$.
Assume $(M,g)$ and $(\tilde M, \tilde g)$ are Riemannian manifolds with an inclusion mapping $\iota: M\hookrightarrow \tilde M$ and $M$ has the submanifold structure.
Suppose $\tilde M$ possesses a shape operator (or any other linear endomorphism) $\tilde s:\tangb \tilde M \to \tangb \tilde M$.
The submanifold structure then gives us an operator $\iota^* \tilde s: \tangb M \to \tangb M$, the pullback of $s$ by $\iota$, via 
\begin{align}
  \forall{X\in\tangb M}\iota^*s X = \dv\iota^{-1} s \dv\iota(X).
\end{align}

\begin{remark}[Remark: Euclidean subspace.]\label{rem:subspace_pullback_shape_operator}
    We consider the special case of the Riemannian manifold $\R^m$.
    Suppose $P:\R^n\to \R^m$ is orthogonal ($n≤m$) and let $s:\R^m\to \R^m$ be a linear operator.
    Then $P^\trsp: \R^n\hookrightarrow \R^m$ is an inclusion map and the \emph{subspace pullback} of $s$ to $\R^n$ is 
    \begin{align}
        \rbr{P^\trsp}^*s = P s P^\trsp. \label{eq:subspace_pullback_shape_operator}
    \end{align}
    We refer to the curvature quantities and directions of $\rbr{P^\trsp}^*s$ with the subspace prefix.
    For example, \emph{subspace principal curvature}.
\end{remark}

\subsubsection{Hypersurfaces}
Let $(M,g)$ be an embedded submanifold of $(\tilde M,\tilde g)$.
Suppose $F:U\subseteq\tilde M \to \R$ is a \emph{local defining function for $M$}, that is, $M\cap U = F^{-1}(\set{0})$ for an open $U\subseteq M$.
Choose arbitrary $X,Y \in \secs( M)$.
From the connection product rule \citep[p.~89]{leeIntroductionRiemannianManifolds2018}
\begin{align}
    \tilde\conn_X\rbr{\textstyle\frac1{\norm{\grad F}}·\grad F  }
    &=  \frac1{\norm{\grad F}}\tilde\conn_X\grad F + \rbr{X \textstyle\frac1{\norm{\grad F}}}·\grad F.
\end{align}
Whence
\begin{align}
    \tilde g\rbr{\tilde\conn_XN  , Y}
        &= \frac{\tilde g\rbr{\tilde\conn_X\grad F, Y}}{\norm{\grad F}} + 
        \rbr{X \frac1{\norm{\grad F}}}\underbrace{\tilde g\rbr{\grad F, Y}}_0 
        \\&=  \frac{\tilde g\rbr{\tilde\conn_X\grad F, Y}}{\norm{\grad F}}.
        \label{eq:covariant_derivative_of_normal}
\end{align}
The zero in the underbrace follows because $\grad F \in \normb M$ and $Y\in\tangb M$.
The Weingarten equation \citep[Thm.~8.13(c)]{leeIntroductionRiemannianManifolds2018} yields $sX = -\tilde\conn_XN$, giving us
\begin{align}
    -\sff_N(X,Y) \overseteqref{eq:shape_operator_def}{=} - g(sX,Y) = \tilde g\rbr{\tilde\conn_XN  , Y} 
    \overseteqref{eq:covariant_derivative_of_normal}{=} \frac{\tilde g\rbr{\tilde\conn_X\grad F, Y}}{\norm{\grad F}}
    \overseteqref{eq:hessian}{=} \frac{\tilde\nabla^2F(X,Y)}{\norm{\grad F}}.\label{eq:second_fundamental_form}
\end{align}

\subsubsection{The Curvature of a Graph Manifold}
\label{sec:graph_manifolds}

The notation and conventions of this section closely follow \citep[\S8]{leeIntroductionRiemannianManifolds2018}.
Let $M$ be the graph of a smooth function $f:\R^n\to\R$, that is, $M = \setcond{(x,f(x))}{x\in \R^n}$.
Since $f$ is smooth $M$ is indeed a manifold \citep[Ex.~1.30, p.~20]{leeIntroductionSmoothManifolds2013}.
We regard $(M,g)$ as a Riemannian submanifold of $(\R^{n+1},\inp{\cdot,\cdot})$, with the inclusion map $\iota: M\hookrightarrow \R^{n+1}$. 

The function $F:\R^{n+1}\to \R$ is a global defining function for $M$ with $F(x,t) \defeq f(x) - t$. Denote by $\nabla f$ the vector of partial derivatives of $f$, and by $\hess_f$ and $\hess_F$, the matrices of second order partial derivatives of $F$ and $f$, all with respect to usual coordinate frames, $(x^i)$, on $\R^n$ and $\R^{n+1}$. Then 
\begin{gather}
  \forall{X,Y\in\tangb \tilde M} \conn^2F(X,Y) = \inp{\hess_F X, Y}.
\end{gather}
At $p\in M$ the differential $\dv \iota_p$ has the Jacobian matrix
\begin{align}
  \begin{pmatrix}
    \id&
    \nabla f(x^1(p),\dots, x^n(p))
  \end{pmatrix}^\trsp.
\end{align}
We identify $\dv\iota_p$ with its Jacobian and $X,Y\in \tangb \tilde M$ with their parameterizations under the coordinate frame, so that we may use the notation of matrix multiplication.

In dropping the $p$-subscripts, the left inverse, $(\dv \iota)^{-1}$, is has the parameterization
\begin{align}
  \rbr{(\dv \iota)^\trsp (\dv \iota)}^{-1}\rbr{\dv \iota}^\trsp 
  =  \rbr{\id + \nabla f\,\nabla f^\trsp}^{-1}\rbr{\dv \iota}^\trsp.
\end{align}
Thus
\begin{align}
  \rbr{\dv \iota}^\trsp \hess_F \rbr{\dv \iota} 
  = \begin{pmatrix}
    \id\\
    \nabla f^\trsp
  \end{pmatrix}^\trsp\begin{pmatrix}
  \hess_F & 0\\ 
  0 & 0
\end{pmatrix}
\begin{pmatrix}
    \id\\
    \nabla f^\trsp
  \end{pmatrix}
  = \hess_f,
\end{align}
whence
\begin{align}
  \rbr{\dv \iota}^{-1} \hess_F \rbr{\dv \iota} 
  &= \rbr{\id + \nabla f\,\nabla f^\trsp}^{-1}\rbr{\dv \iota}^\trsp\hess_F \rbr{\dv \iota} 
  \\&= \rbr{\id + \nabla f\,\nabla f^\trsp}^{-1}\hess_f.\label{eq:inclusion_pseudo_inverse}
\end{align}

We can now compute the shape operator of $M$ associated to the unit normal field $\grad F/\norm{\grad F}$.
Pick arbitrary $X,Y\in\tangb M$. From \eqref{eq:second_fundamental_form} there is 
\begin{align}
  -\sff_N(X,Y) · \norm{ \grad F}
  &\overseteqref{eq:second_fundamental_form}{=} \tilde g(\tilde \conn_X\grad F, \dv\iota(Y))
  \\&= g\rbr{\dv\iota^{-1}\circ\hess_F \circ\dv\iota (X), Y},\label{eq:scalar_second_fundamental}
\end{align}
and, using the bilinearity of of $g$ we have 
\begin{gather}
  g\rbr{sX ,Y}\overseteqref{eq:shape_operator_def}{=} \sff_N(X,Y) \overseteqref{eq:scalar_second_fundamental}{=}  - \frac{g(\dv\iota^{-1} \hess_F \dv\iota X, Y)}{\norm{\grad F}}.
  \intertext{This, together with \eqref{eq:inclusion_pseudo_inverse} yields the shape operator:}
  s = -\frac{\rbr{\id + \nabla f\;\nabla f^\trsp}^{-1}\hess_f}{
    \sqrt{\norm{\nabla f}^2 + 1}
  }.\label{eq:shape_operator}
\end{gather}

\subsubsection{The Curvature of a Level-Set Manifold}\label{sec:level_set_method}
We now consider the case where $M$ is the level-set of a smooth function $f:\R^n \to \R$ so that for some $c\in\R$ we have $M = \setcond{ x \in \R^n }{f(x) = c}$.  The following theorem can be used to ensure the level-sets of $f$ do indeed posses a manifold structure, which is a simplification of the constant rank level-set theorem \citep[Thm.~5.12]{leeIntroductionSmoothManifolds2013} for our setting.

\begin{theorem}[Theorem.]
  Let $f:\R^n\to\R$ be a smooth function having constant rank $r$. Then the level-sets of $f$ are all smooth manifolds of codimension $r$.
\end{theorem} 

We use the same construction as \citet[Ex.~1.32, \S5]{leeIntroductionSmoothManifolds2013} to calculate the Gaussian and mean curvature at a point $(x_0,y_0)\in (\R^{n-1}\times \R) \cap M$. 

From the implicit function theorem \citep[Thm.~C.40]{leeIntroductionSmoothManifolds2013} there are open connected neighborhoods $U_0\times V_0 \subseteq \R^{n-1}\times \R$ with $ (x_0,y_0)\in U_0\times V_0$ and a smooth function $g:U_0\to V_0$ so that $(U_0\times V_0) \cap M$ is the graph of $g$. That is, $(U_0\times V_0) \cap M = \setcond{(x,g(x)) }{ x \in U_0 \subseteq\R^{n-1}}$. Consequentially the gradient of $g$ has the structure 
\begin{align}
  \forall{x \in U_0} f(x, g(x)) = c 
  &\implies \Dv_1 f(x, g(x)) + \Dv_2 f(x, g(x)) \nabla g(x) = 0
  \\&
  \iff     \nabla g(x) = -\sbr{\Dv_2 f}^{-1}(x, g(x)) \Dv_1 f(x, g(x)).
  \label{eq:implicit_function_theorem}
\end{align}
Let $\partial_i \defeq \diffp{}/{x^j}$ and $\partial_{ij}^2 \defeq \diffp{}/{x^i,x^j}$. Then for  $i,j \in[n-1]$ and $x\in U_0$ 
\begin{gather}
  \partial_i g(x) = -\rbr{\partial_n f(x, g(x))}^{-1}\partial_i f(x, g(x)), \label{eq:grad_verbose}
\end{gather}
and
\begin{align}
  \partial_{ij}^2 g(x)
  &= \frac1{\rbr{ \partial_n f}^{2}}\sbr2{
    \partial_{nj}^2 f 
    + \partial^2_{nn}f \partial_j g
    }\rbr{\partial_i f}
    -  \frac1{ \partial_n f}\sbr2{\partial^2_{ij}f + \partial_{nj}^2 f \partial_j g   }
  \\&= -\frac1{ \partial_n f}\sbr2{
    \partial_{nj}^2 f 
    + \partial^2_{nn}f\partial_j g
    }\rbr{\partial_i g}
    -  \frac1{ \partial_n f}\sbr2{\partial^2_{ij}f + \partial_{nj}^2 f \partial_j g   }
  \\&= -\frac1{ \partial_n f}\sbr2{
        \partial_{nj}^2 f \partial_i g
        + \partial^2_{nn}f \partial_i g\partial_j g
        + \partial^2_{ij}f + \partial_{nj}^2 f \partial_j g  
        }
  \\&= -\frac1{ \partial_n f}\sbr2{
        \partial_{nj}^2 f \rbr{\partial_i g + \partial_j g}
        + \partial^2_{nn}f \partial_i g\partial_j g
        + \partial^2_{ij}f 
     },\label{eq:hess_verbose}
\end{align}
where every function on the right hand side is evaluated at either the point $x$ or $(x,g(x))$ (depending on the function).\footnote{This is easy to do at  $x_0$ since $(x_0, g(x_0))= (x_0, y_0)$.} If we let $\bar\nabla \defeq \rbr*{\diffp{}/{x^1}, \dots, \diffp{}/{x^{n-1}}}^\trsp$, and $\bar\hess$ be, similarly, the $(n-1)\times(n-1)$ matrix of cross-partial derivatives, there are the more compact expressions for \eqref{eq:grad_verbose} and \eqref{eq:hess_verbose}:
\begin{gather}
  \nabla g = -\frac1{\partial_n f} \bar\nabla f,
  \intertext{and}
  \hess_g = -\frac1{\partial_n f}\sbr2{
    \rbr{\nabla g + \nabla g^\trsp}  \diag\rbr{\bar\nabla \partial_n f }
    + \partial^2_{nn} f · \rbr{\nabla g \nabla g^\trsp} 
    + \bar \hess_f 
  },
\end{gather}
where the sum $\nabla g + \nabla g^\trsp$ is to be interpreted as the matrix of pairwise sums of the elements of $\nabla g $: $\rbr{\nabla g + \nabla g^\trsp}_{ij} \defeq \diffp{g}/{x^i} + \diffp{g}/{x^j}$.
Then, as in \autoref{sec:graph_manifolds}, $F(x) \defeq g(x^1,\dots,x^{n-1}) - x^n$ is a local defining function for $M$ on $(U_0\times V_0)\cap M$. The local graph parameterization provided by $g$ is $\phi:U_0\to M$ with $\phi(x) \defeq (x,g(x))$, and its differential has the parameterization
\begin{gather}
  \dv\phi = \begin{pmatrix}
    \id 
    & \nabla g
  \end{pmatrix}^\trsp.
\end{gather}
Finally, \textit{mutatis mutandis}, the shape operator is given by \eqref{eq:shape_operator}.

\begin{remark}[Numerical considerations]
    \label{rem:change_of_basis}
    In order to compute the level-set curvature we needed to apply the implicit function theorem.
    However, it may be difficult numerically to stabilize the inverse operation in \eqref{eq:implicit_function_theorem}.
    To remedy this problem, one may apply a change of basis before computing the quantities \eqref{eq:grad_verbose} and \eqref{eq:hess_verbose}.
    Let $B\in\R^{n \times n }$ be an orthogonal change of basis operator.
    Then we apply all our operations to $B^\trsp M \defeq \setcond{ B^\trsp x}{x \in M}$.
    
    Differentiating $\tilde f \defeq f\circ B^\trsp$ yields the revised quantities
    \begin{gather}
        \forall{x\in\R^n}\nabla \tilde f(x) = B · \nabla f(B^\trsp x) 
        \quad\text{and}\quad
        \hess_{\tilde f}\vert_{x} = B \hess_f\vert_{(B^\trsp x)} B^\trsp.
    \end{gather}
\end{remark}

\subsubsection{Summary}
In order to study the curvature over a given subspace we introduced the \emph{subspace pullback} of $s$
\begin{gather}\label{eq:subspace_pullback}
    \rbr{P^\trsp}^*s \defeq P s P^\trsp,
\end{gather}\noeqref{eq:subspace_pullback}
where $P\in\R^{m \times n}$ is an orthogonal projection, so that the image of $P^\trsp$ spans the subspace of $\R^m$ in question.
This definition of $\rbr{P^\trsp}^*s$ essentially restricts the shape operator to a linear subspace of the tangential space and is motivated by the principles of differential geometry.
The pullback shape operator allows us to compare the amount of curvature in the subspace to that of the entire response space.

\section{Dataset and models}\label{app:datasets_models}
\setlength{\tabcolsep}{2pt}
\begin{table}
\caption{Accuracies of CIFAR models at different perturbation lengths}
\label{tab:cifar_accs}
\resizebox{\textwidth}{!}{%
    \begin{tabular}{l|llllllllll}
       \toprule
       & \multicolumn{10}{c}{Dimension} \\
       $\epsilon$   & 1 & 2 & 3 & 4 & 5 & 6 & 7 & 8 & 9 & 10 \\
       \midrule
    0   & 95.25\% / & 95.25\% / & 95.25\% / & 95.25\% / & 95.25\% / & 95.25\% / & 95.25\% / & 95.25\% / & 95.25\% / & 95.25\% / \\
        & 90.83\% & 90.83\% & 90.83\% & 90.83\% & 90.83\% & 90.83\% & 90.83\% & 90.83\% & 90.83\% & 90.83\% \\[.2cm]
    0.1 &51.43\% / & 62.86\% / & 70.48\% / & 76.2\% / & 79.06\% / & 80.01\% / & 81.92\% / & 81.92\% / & 82.87\% / & 83.82\% / \\
        &86.29\% & 90.83\% & 90.83\% & 90.83\% & 90.83\% & 90.83\% & 90.83\% & 90.83\% & 90.83\% & 90.83\% \\[.2cm]
    0.25&1.91\% / & 4.76\% / & 5.72\% / & 6.67\% / & 7.62\% / & 8.57\% / & 9.52\% / & 9.52\% / & 9.52\% / & 10.48\% / \\
        &79.93\% & 89.01\% & 89.92\% & 90.83\% & 90.83\% & 90.83\% & 90.83\% & 90.83\% & 90.83\% & 90.83\% \\[.2cm]
    0.5 &0.0\% / & 0.0\% / & 0.0\% / & 0.0\% / & 0.0\% / & 0.0\% / & 0.0\% / & 0.0\% / & 0.0\% / & 0.0\% / \\
        &65.4\% & 84.47\% & 86.29\% & 87.2\% & 87.2\% & 87.2\% & 87.2\% & 87.2\% & 87.2\% & 87.2\% \\[.2cm]
    1   &0.0\% / & 0.0\% / & 0.0\% / & 0.0\% / & 0.0\% / & 0.0\% / & 0.0\% / & 0.0\% / & 0.0\% / & 0.0\% / \\
        &36.33\% & 54.5\% & 59.04\% & 61.76\% & 65.4\% & 69.03\% & 69.94\% & 70.85\% & 72.66\% & 73.57\% \\[.2cm]
    2   &0.0\% / & 0.0\% / & 0.0\% / & 0.0\% / & 0.0\% / & 0.0\% / & 0.0\% / & 0.0\% / & 0.0\% / & 0.0\% / \\
        &2.72\% & 5.45\% & 10.9\% & 12.72\% & 13.62\% & 14.53\% & 15.44\% & 15.44\% & 16.35\% & 17.26\% \\[.2cm]
    \end{tabular}}
\end{table}

For the adversarial subspace analysis we used two identical ResNet-50 architectures \citep{he2016deep, engstrom2019robustness}\footnote{Pretrained models were obtained from \url{https://github.com/MadryLab/robustness}} and two identical WideResNet-70 architectures \citep{gowal2020uncovering}\footnote{Pretrained models were obtained from \url{https://github.com/deepmind/deepmind-research/tree/master/adversarial_robustness}}.
For each architecture, one model was trained on the unmodified CIFAR-10 dataset \citep{krizhevsky2009learning}.
For the second model we additionally incorporated adversarial examples in the training process, widely known as adversarial training.
A training perturbation size $\epsilon$ of 0.5 was used. 
To perform the curvature analysis, we used twice-differentiable ELU activation functions \citep{clevert2015fast} for both models, which had little impact on the clean or adversarial accuracies.

In the analysis we include 100 CIFAR-10 test images, class-balanced, all unseen during training, and correctly classified by both models.
The optimization algorithm stopped after finding 50 orthogonal adversarial vectors for all 100 test images.
For the curvature analysis we used a subset of 50 class-balanced images and 10 adversarial directions per image.

Additionally, we only considered one seed per CIFAR-10 model because we found that the analysis on the MNIST trained models had little variation across seeds (see Appendix \ref{app:seeds}).
The accuracy of both models at different perturbation lengths $\epsilon$ and dimensions is shown in Table \ref{tab:cifar_accs}.
Here, the first line of each row corresponds to the accuracy of the naturally trained model, whereas the second line is referring to the adversarially trained model.

We repeated many of the experiments on the MNIST dataset \citep{lecun1998gradient} and found largely consistent results (see Appendix \ref{app:mnist}).

\section{Additional CIFAR-10 results}\label{app:add_cifar}
\begin{figure}
  \centering
  \includegraphics[width=\textwidth]{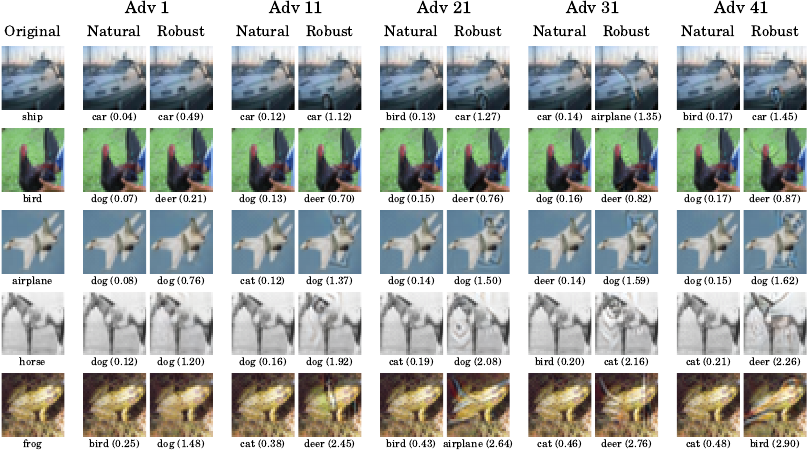}
  \caption{\textbf{CIFAR-10 example adversarials}. The panels show adversarial examples for a given input image. Column pairs are sorted by adversarial dimension. Within a pair, each column includes examples for the natural and robust models, respectively. The labels below the images indicate the class assigned by the model. The numbers in brackets are the $\ell_2$-norms of the respective images.}
  \label{fig:cifar_examples}
  \vspace{-5pt}
\end{figure}

\begin{wrapfigure}{R}{0.60\textwidth}
 \vspace{-15pt}
 \centering
 \includegraphics[scale=1.00]{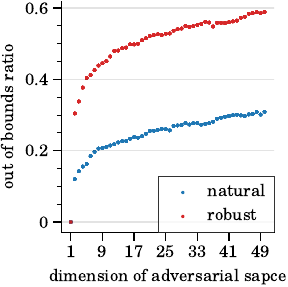}
 \caption{\textbf{Ratio of out of bounds samples}.
 We uniformly sample the distance to the decision boundary with increasing dimensionality per input image and adversarial subspace dimension.
 We report the fraction of directions pointing to out-of-bounds boundaries, averaged across 100 images.
 }
 \label{fig:cifar_out_bnd}
 \vspace{-5pt}
\end{wrapfigure}

Here we show additional analysis to complement our findings on the CIFAR-10 dataset.

Figure \ref{fig:cifar_examples} shows five adversarially perturbed images and their corresponding original image found for the two ResNet-50 models.
We can see that that the perturbation length is increasing with dimension number.
Interestingly, while the adversarial categories may not be semantically adjacent to the clean classes, the adversarially trained model seems to require more semantically meaningful perturbations, e.g. the images of boats in the top row, misclassified as cars, are perturbed to have elements looking like wheels.
We can also see that there is no clear, consistent $\ell_2$-length of an adversarial attack that makes an image perturbation ambiguous for a human observer.
For example, the robust model example in the first (ship) row of the adversarial 11 column pair has a perturbation length of 1.12 and could be mistaken as a car on a first glance.
In contrast, the robust model example in the fourth (horse) row of the adversarial 1 column pair is still clearly identifiable as a horse with a larger perturbation length of 1.2.
This illustrates that the direction of the perturbation is as important as the perturbation magnitude itself, as was previously reported by \citep{goodfellow2014explaining, ilyas2019adversarial}, although assessing the ``perceptual distance'' is an open area of research.

As was described earlier, we uniformly randomly sampled directions within adversarial subspaces to measure the distance to the decision boundary.
Figure \ref{fig:cifar_out_bnd} shows how many of the samples pointed to decision boundaries that were outside of the allowed input pixel bounds.

\begin{figure}
  \centering
  \includegraphics[width=\textwidth]{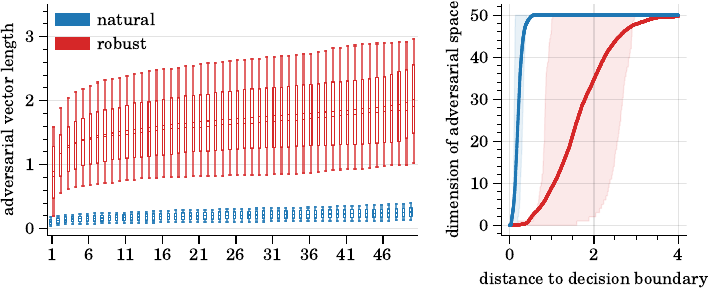}
  \caption{\textbf{Perturbation lengths and dimensionality}. (left) The subfigure shows the distribution of the n-th found adversarial perturbations across images. (right) We tested at distance increments of 0.001 how many adversarial dimensions were found at that distance per image. The lines represent the means over all test images and the color shaded areas show the 10th and 90th percentiles. }
  \label{fig:pert_lengths}
\end{figure}

Figure \ref{fig:pert_lengths} shows that perturbation lengths are increasing sublinearly and monotonically with increasing number of adversarial dimensions.
For the naturally trained model there is also a much larger increase of adversarial dimension with increasing distance to distance to decision boundary.
This figure is different from e.g. Figure \ref{fig:cifar_perturbation_distances} in that it displays the lengths of the adversarial subspace axes, not the lengths of the random perturbations within these subspaces.
Note that in the right panel of this figure both models plateau at 50 dimensions because we stopped the optimizer at this point.

\begin{figure}
  \centering
  \includegraphics[width=\textwidth]{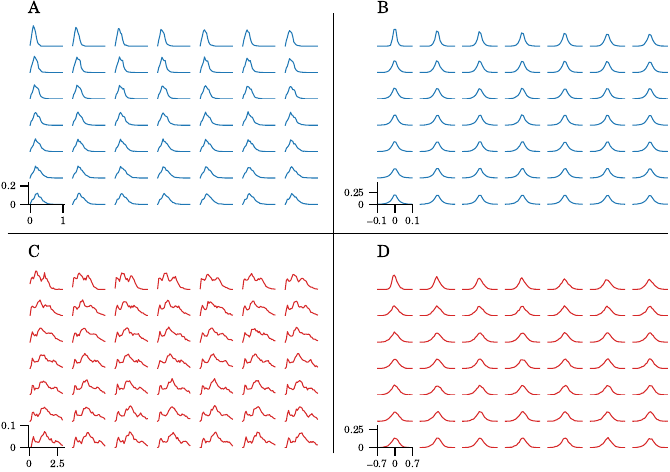}
  \caption{\textbf{Distributions of distances to decision boundary}.
  (A,B) Distributions for natural model.
  (C,D) Distributions for adversarially trained model.
  (A-C) We show the complete set histograms corresponding to Figure \ref{fig:cifar_perturbation_distances} dimensions.
  (C,D) Additionally, we provide the mean-normalized histograms to corresponding to the histograms in A and C to emphasize the smooth increase in variance of distances with increasing dimensionality.
  }
  \label{fig:cifar_distributions}
\end{figure}

We also provide additional, and more detailed information on the distributions of distances to the decision boundary of randomly sampled vectors in adversarial subspaces in Figure \ref{fig:cifar_distributions}.
Each subplot shows a subspace of increasing dimensionality, ascending left-to-right, then top-to-bottom.
We observe that the variance of the distances to the decision boundary of randomly sampled vectors is smoothly increasing with dimensionality.

\subsubsection*{Additional model comparison}

\begin{figure}[t!]
  \centering
  \includegraphics[width=\textwidth]{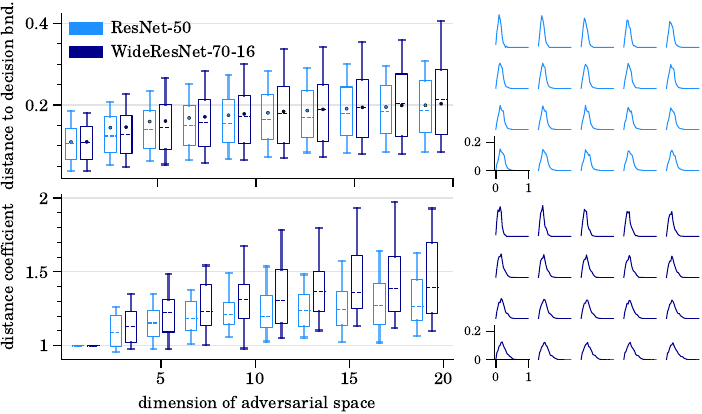}
  \caption{\textbf{Distances to the decision boundary}. Compare with Figure \ref{fig:cifar_perturbation_distances}.
  }
  \label{fig:cifar_perturbation_distances_comp_nat}
  \vspace{-1pt}
\end{figure}

\begin{figure}
  \centering
  \includegraphics[width=\textwidth]{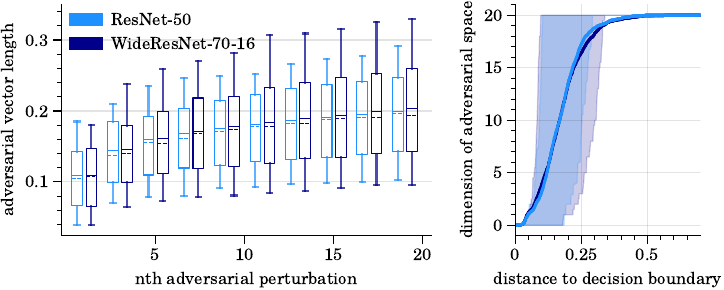}
  \caption{\textbf{Perturbation lengths and dimensionality}. Compare with Figure \ref{fig:pert_lengths}.
  }
  \label{fig:pert_lengths_comp_nat}
\end{figure}

\begin{figure}[t!]
  \centering
  \includegraphics[width=\textwidth]{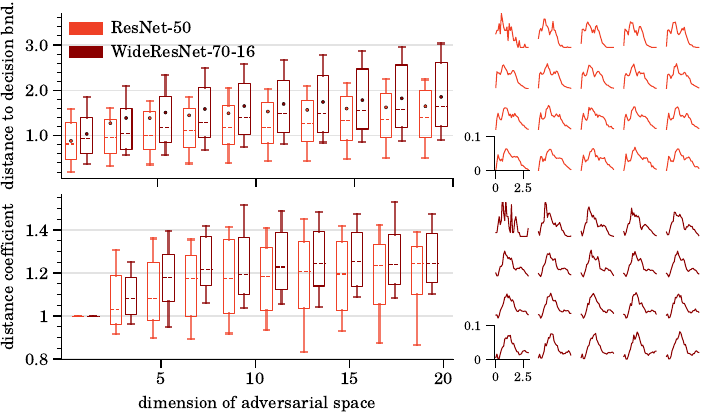}
  \caption{\textbf{Distances to the decision boundary}. Compare with Figure \ref{fig:cifar_perturbation_distances}.
  }
  \label{fig:cifar_perturbation_distances_comp_rob}
  \vspace{-1pt}
\end{figure}

\begin{figure}
  \centering
  \includegraphics[width=\textwidth]{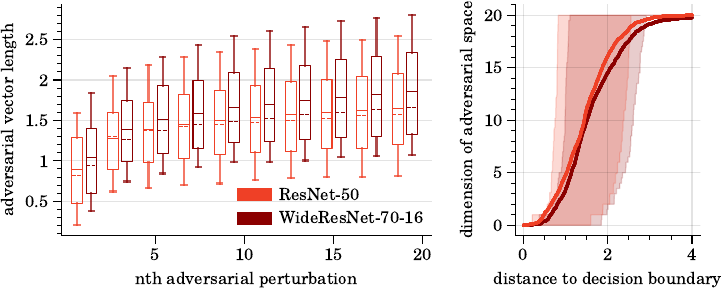}
  \caption{\textbf{Perturbation lengths and dimensionality}. Compare with Figure \ref{fig:pert_lengths}.
  }
  \label{fig:pert_lengths_comp_rob}
\end{figure}

\begin{figure}
\centering
\begin{subfigure}{0.5\textwidth}
 \centering
 \includegraphics{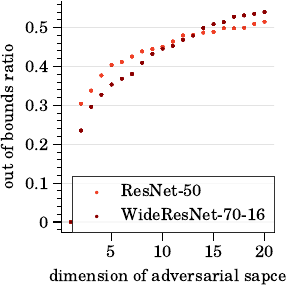}
 \caption{}
 \label{fig:cifar_out_bnd_comp_rob}
\end{subfigure}%
\begin{subfigure}{.5\textwidth}
 \centering
 \includegraphics{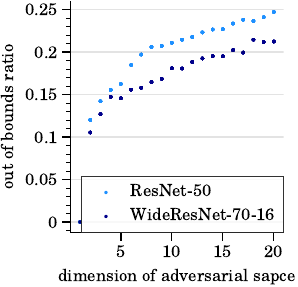}
  \caption{}
 \label{fig:cifar_out_bnd_comp_nat}
\end{subfigure}
 \caption{\textbf{Ratio of out of bounds samples}. Compare with Figure \ref{fig:cifar_out_bnd}.
 }
\label{fig:cifar_out_bnd_comp}
\end{figure}

To show that our findings hold for different models, we present results for two different architectures: a ResNet-50 \citep{he2016deep} and a WideResNet-70-16, of which the latter is the state-of-the-art adversarially trained network for CIFAR-10 images \citep{gowal2020uncovering}.
Both architectures were trained with and without adversarial perturbations, resulting in 4 models overall.
Figures \ref{fig:cifar_perturbation_distances_comp_nat} through \ref{fig:cifar_out_bnd_comp} repeat analysis to compare  natural and adversarial trained models, but instead compare models of different architectures and constant training procedure.

Figures \ref{fig:cifar_perturbation_distances_comp_nat}, \ref{fig:pert_lengths_comp_nat}, and \ref{fig:cifar_out_bnd_comp_nat} show a comparison of the two naturally trained models.
These figures indicate that there is little difference between the two model architectures regarding distance to the decision boundary and perturbation length over dimensions.
Figures \ref{fig:cifar_perturbation_distances_comp_rob}, \ref{fig:pert_lengths_comp_rob}, and \ref{fig:cifar_out_bnd_comp_rob} show a comparison of the two adversarially trained models.
Here, the figures show that perturbation lengths as well as distances to the decision boundary are larger for the WideResNet-70-16 compared to the ResNet-50. This is expected, as the WideResNet has higher adversarial accuracy and thus larger perturbations are needed to fool it.
At the level of these aggregate statistics, the hyperparameter differences in adversarial training (Figure \ref{fig:cifar_perturbation_distances_comp_rob}) have more of an effect than the model architecture differences themselves (Figure \ref{fig:cifar_perturbation_distances_comp_nat}), although all differences are much less pronounced than the differences between adversarially and naturally trained networks (Figure \ref{fig:cifar_perturbation_distances}).

\begin{figure}[ht]
  \centering
  \includegraphics[width=\textwidth]{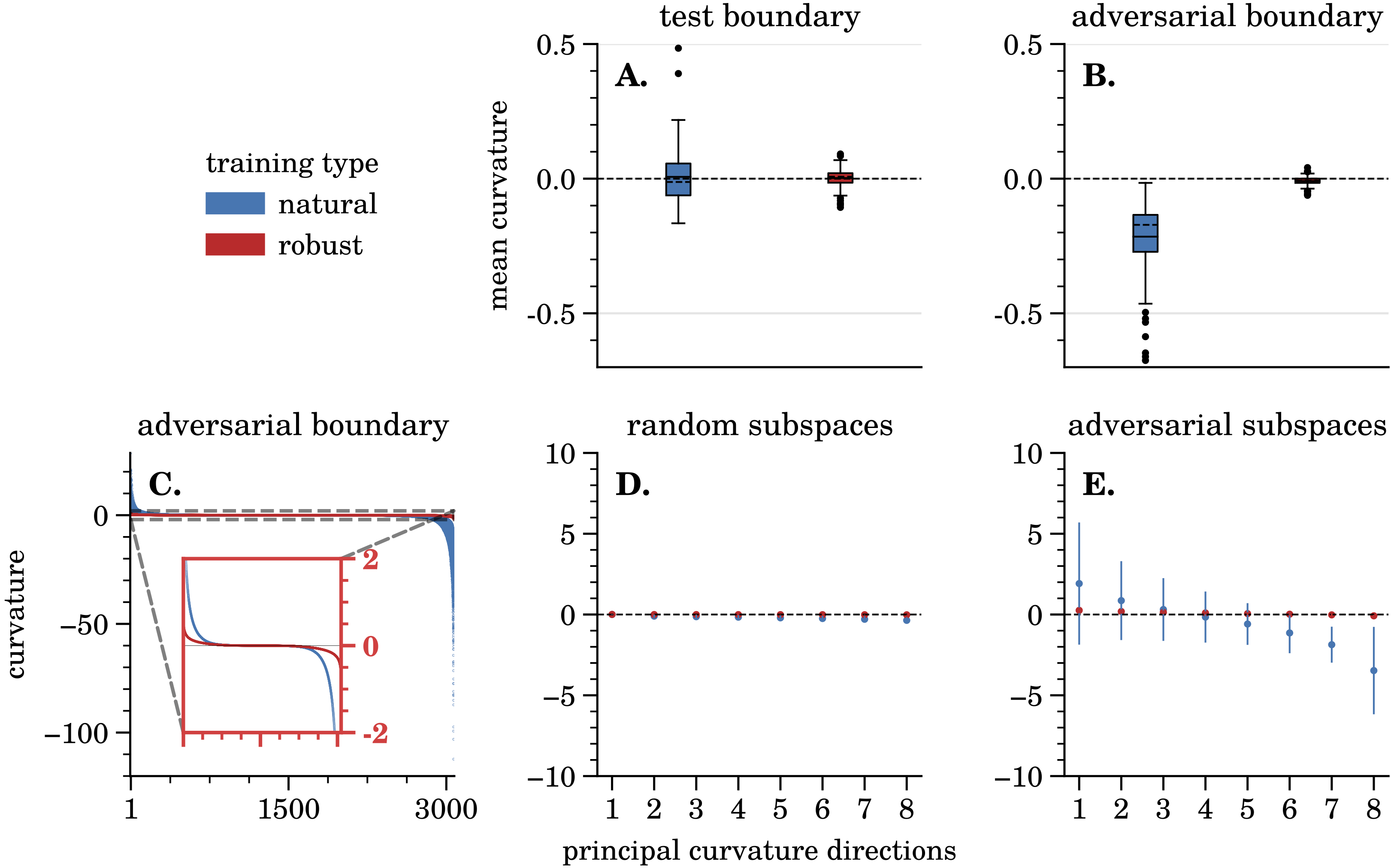}
  \caption{\textbf{Decision boundary curvature.}
  WRN-70 curvature analysis. Compare with Figure \ref{fig:curvature_analysis}.
  }
  \label{fig:wrn_curvature_analysis}
  \vspace{-8pt}
\end{figure}

\begin{figure}[ht]
  \centering
  \includegraphics{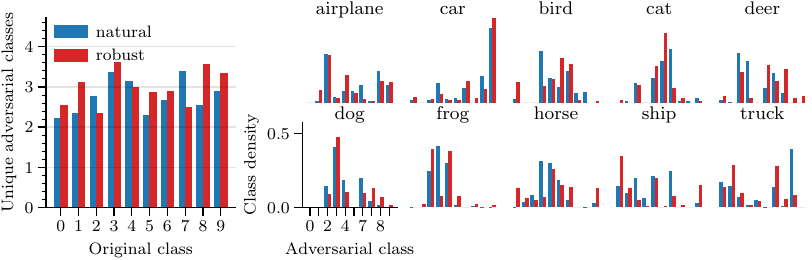}
  \caption{\textbf{Adversarial Class Composition.}
  WRN-70 adversarial class analysis. Compare with Figure \ref{fig:adv_class_composition}.
  }
  \label{fig:wrn_adv_class_composition}
  \vspace{-8pt}
\end{figure}

\begin{wraptable}{r}{0.4\textwidth}
   \centering
   \vspace{-10pt}
   \begin{tabular}{c c c}
   \\
   \toprule
   Label & Natural & Robust \\
   \midrule
   airplane & 1.83 & \bf{1.84} \\
   car & \bf{1.47} & 1.44 \\
   bird & \bf{1.69} & 1.63 \\
   cat & \bf{1.53} & 1.41 \\
   deer & 1.55 & \bf{1.80} \\
   dog & 1.49 & \bf{1.61} \\
   frog & 1.23 & \bf{1.38} \\
   horse & 1.65 & \bf{1.95} \\
   ship & \bf{1.78} & 1.77 \\
   truck & 1.68 & \bf{1.79} \\
   \bottomrule
   \end{tabular}
   \caption{Entropy of unique proximal adversarial labels per class, averaged across a balanced sampling of correctly labeled test images. For 6 out of 10 categories the entropy is increased with adversarial training. Results are for the WRN-70 network. Compare with Table \ref{table:class_entropy}.}
   \vspace{-10pt}
   \label{table:wrn_class_entropy}
\end{wraptable}
The box plots depicting the distances to the decision boundary in Figures \ref{fig:cifar_perturbation_distances_comp_nat} and \ref{fig:cifar_perturbation_distances_comp_rob} show that, for both architectures, the adversarial subspace is large, continuous, and increases sublinearly in volume as the dimensionality is increased.
The observation that this finding is agnostic to model architecture is further supported by the similarity in distributions of samples measuring the distance to decision boundary that can be seen on the right hand side of the same figures.
For Figure \ref{fig:wrn_curvature_analysis} we repeated the curvature analysis on the natural and robust neural networks, although we limited the experiment to 10 images and 8 dimensional adversarial subspaces due to the increase in computational cost associated with the larger model size.
Comparing this figure against Figure \ref{fig:curvature_analysis} shows that the mean and principal curvature profiles are largely consistent.
Finally, Table \ref{table:wrn_class_entropy} and Figure \ref{fig:wrn_adv_class_composition} demonstrate that adversarial training results in an additional diversity of nearby classes within adversarial subspaces for the WRN-70 network, although the effect is less pronounced than for the RN-50 network.
Thus, the observed trends and conclusions drawn from the analysis provided in this study hold across architectures.

\section{Additional MNIST results}\label{app:mnist}
We conducted the same experiments described in Section \ref{sec:results} on networks trained with the MNIST dataset.
As a test set we used 500 images (50 per class), unseen in training and correctly labeled by all investigated models.

In Figure \ref{fig:mnist_pert_lengths} we can see that perturbation lengths are much larger for MNIST than for CIFAR-10, although the relative increase in distances for the robust network is still present.
The optimization was in general more difficult with MNIST, likely due to the fact that the images all live very close to the allowable input bounds.
Specifically, the algorithm found less than 30 adversarial dimensions for more than 95\% of the test images (Figure \ref{fig:optimizer_seeds}), to be compared with finding >50 dimensions for 100\% of the CIFAR-10 test images.

\begin{figure}[ht]
 \vspace{-5pt}
 \centering
 \includegraphics{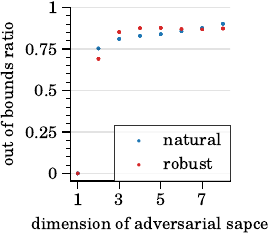}
 \caption{\textbf{Ratio of out of bounds samples}.
 Compare to Figure \ref{fig:cifar_out_bnd}.}
  \label{fig:mnist_out_bnd}
 \vspace{-5pt}
\end{figure}
That it is harder to fool the MNIST model can also be seen in Table \ref{tab:mnist_accs}: the accuracies of both models are much higher for the same perturbation lengths than the accuracies of the CIFAR-10 models.
For the curvature analysis, on the other hand, we found that the overall values were larger on CIFAR-10.
Additionally, the adversarially trained model had more decision boundary curvature within adversarial subspaces on MNIST than on CIFAR-10.

\setlength{\tabcolsep}{2pt}
\begin{table}
\caption{Accuracies of MNIST models (seed 0) at different perturbation lengths}
\label{tab:mnist_accs}
\resizebox{\textwidth}{!}{%
    \begin{tabular}{l|llllllllll}
       \toprule
        & \multicolumn{10}{c}{Dimension} \\
       $\epsilon$   & 1 & 2 & 3 & 4 & 5 & 6 & 7 & 8 & 9 & 10 \\
       \midrule
        0 & 98.59\% / & 98.59\% / & 98.59\% / & 98.59\% / & 98.59\% / & 98.59\% / & 98.59\% / & 98.59\% / & 98.59\% / & 98.59\% / \\
         & 98.56\% & 98.56\% & 98.56\% & 98.56\% & 98.56\% & 98.56\% & 98.56\% & 98.56\% & 98.56\% & 98.56\% \\[.2cm]
        0.5 & 90.51\% / & 97.41\% / & 98.0\% / & 98.39\% / & 98.39\% / & 98.39\% / & 98.37\% / & 98.35\% / & 98.29\% / & 98.23\% / \\
         & 97.77\% & 98.56\% & 98.56\% & 98.56\% & 98.56\% & 98.56\% & 98.56\% & 98.56\% & 98.56\% & 98.56\% \\[.2cm]
        1 & 60.53\% / & 90.31\% / & 95.83\% / & 97.01\% / & 97.4\% / & 97.58\% / & 97.94\% / & 97.86\% / & 97.69\% / & 97.51\% / \\
         & 94.03\% & 98.36\% & 98.56\% & 98.56\% & 98.56\% & 98.56\% & 98.56\% & 98.56\% & 98.56\% & 98.56\% \\[.2cm]
        2 & 1.77\% / & 18.34\% / & 47.91\% / & 71.58\% / & 83.54\% / & 88.51\% / & 91.63\% / & 93.0\% / & 93.8\% / & 93.55\% / \\
         & 68.99\% & 89.3\% & 95.01\% & 96.59\% & 97.57\% & 97.69\% & 97.73\% & 97.72\% & 98.56\% & 98.56\% \\[.2cm]
        4 & 0.0\% / & 0.0\% / & 0.2\% / & 4.14\% / & 15.05\% / & 26.41\% / & 32.43\% / & 38.61\% / & 37.46\% / & 32.74\% / \\
         & 0.59\% & 9.46\% & 27.99\% & 46.52\% & 65.77\% & 76.32\% & 82.59\% & 83.94\% & 85.42\% & 80.16\% \\[.2cm]
        6 & 0.0\% / & 0.0\% / & 0.0\% / & 0.39\% / & 1.98\% / & 7.86\% / & 10.45\% / & 18.94\% / & 16.48\% / & 16.55\% / \\
         & 0.0\% & 0.0\% & 0.2\% & 7.69\% & 24.69\% & 37.51\% & 47.63\% & 48.86\% & 54.54\% & 40.74\% \\[.2cm]
    \end{tabular}}
\end{table}

We noticed that most of the MNIST adversarials lie in a thin regions between the decision boundary and the valid pixel box boundary (compare Figure \ref{fig:mnist_boundary_visualizations} against Figure \ref{fig:boundary_visualizations}).
Figure \ref{fig:mnist_out_bnd} shows that this leads to large rates of out of bounds samples when measuring the distance to the decision boundary.
For example, 90\% of the time the decision boundary is outside the valid pixel range in the 6 dimensional adversarial subspace of the naturally trained model.

\begin{figure}[ht]
  \centering
  \includegraphics[width=\textwidth]{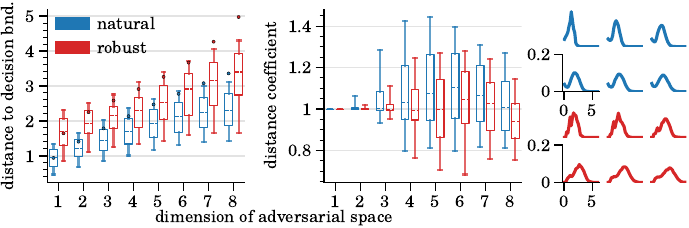}
  \caption{\textbf{MNIST distances to the Decision Boundary.}
    We repeated the distance analysis from Figure \ref{fig:cifar_perturbation_distances} on the MNIST dataset
    Only test images with at least 8 adversarial dimensions were included in this Figure.
  }
\label{fig:mnist_perturbation_distances}
\end{figure}

\begin{figure}[ht]
  \centering
  \includegraphics[width=\textwidth]{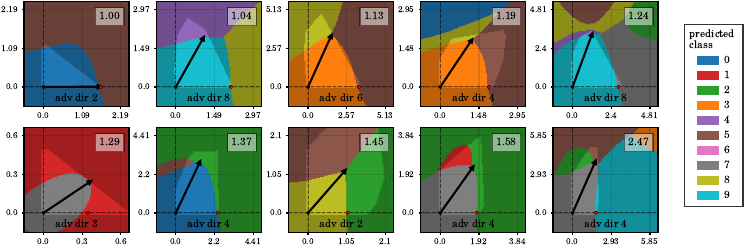}
  \caption{\textbf{MNIST boundary visualizations.}
  This figure is the MNIST version of Figure \ref{fig:boundary_visualizations}.
  Note that for MNIST, the decision boundary lies outside the valid pixel range more often than for CIFAR.
  }
  \label{fig:mnist_boundary_visualizations}
\end{figure}

\begin{figure}[ht]
  \centering
  \includegraphics[width=\textwidth]{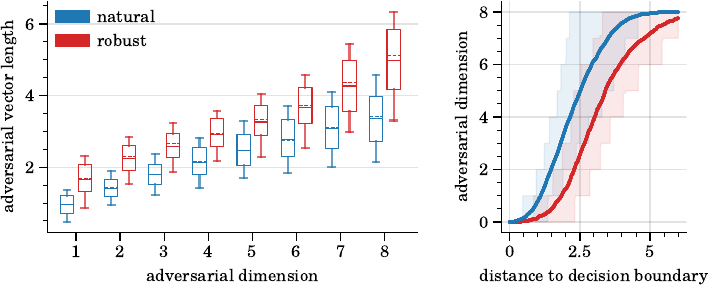}
  \caption{\textbf{MNIST perturbation lengths and dimensions.}
    Compare with Figure \ref{fig:pert_lengths}.
  }
  \label{fig:mnist_pert_lengths}
\end{figure}

\begin{figure}[ht]
  \centering
  \includegraphics[width=\textwidth]{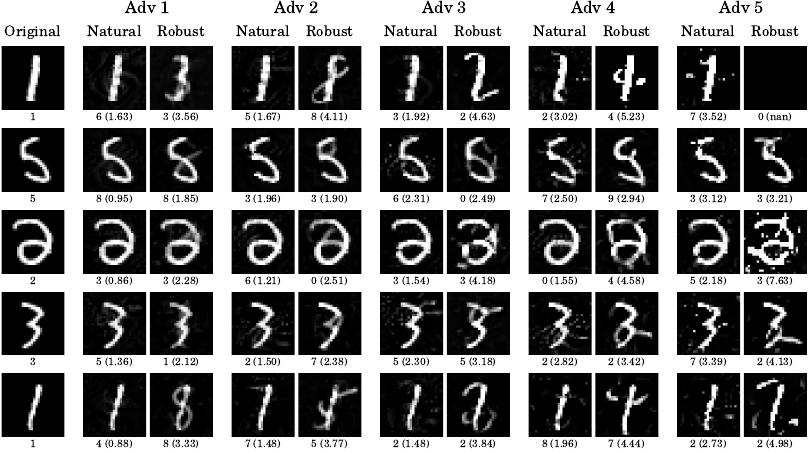}
  \caption{\textbf{MNIST example adversarials.}
    Compare to Figure \ref{fig:cifar_examples}.
  }
  \label{fig:samples_MNIST}
\end{figure}

\begin{figure}[ht]
  \centering
  \includegraphics[width=\textwidth]{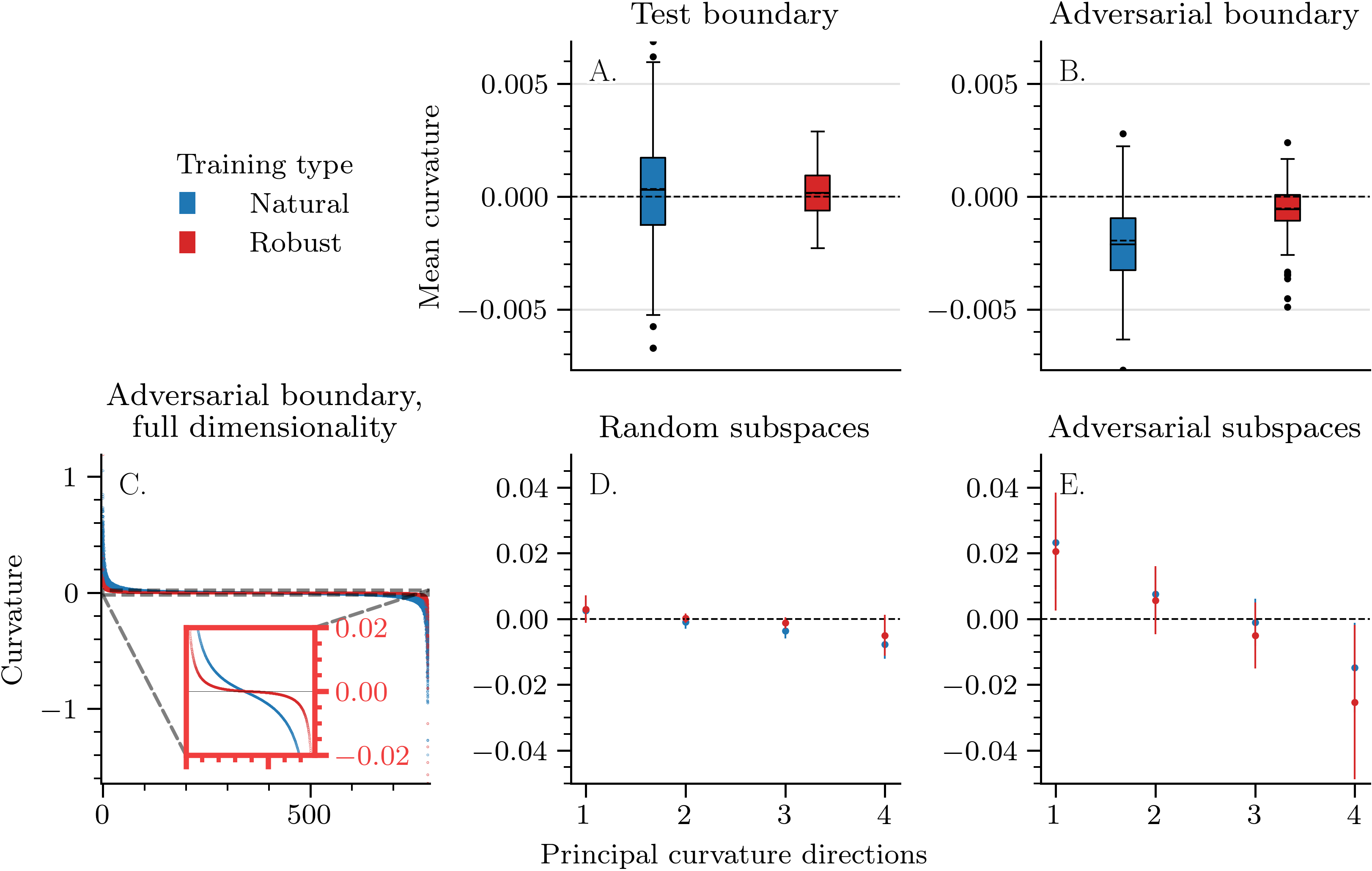}
  \caption{\textbf{MNIST decision boundary curvature.}
  Compare with Figure \ref{fig:curvature_analysis}
  }
  \label{fig:mnist_curvature_analysis}
\end{figure}

\section{Seed consistency}\label{app:seeds}
\begin{wrapfigure}{R}{0.5\textwidth}
 \vspace{-5pt}
 \centering
  \includegraphics{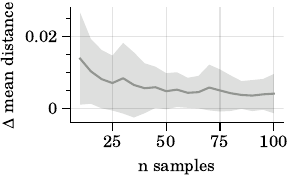}
  \caption{\textbf{Sampling size in adversarial subspaces}.
  The line shows the mean running difference in distance to the decision boundary means across 100 samples as one increases the sample size from 5 to 100. The shaded area depicts the standard deviation from the mean line.}
  \label{fig:sample_convergence}
 \vspace{-5pt}
\end{wrapfigure}

\begin{figure}[ht]
  \centering
  \includegraphics{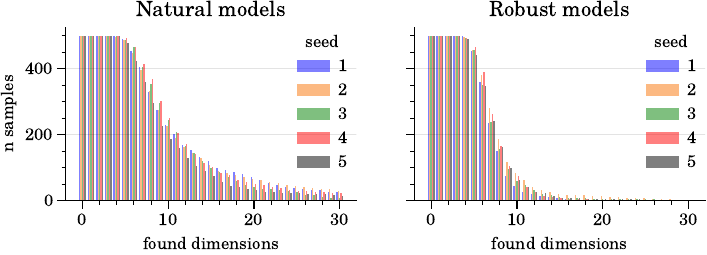}
  \caption{\textbf{Optimization consistency on MNIST}.
  The bars show number of samples for which at least n dimensions were found during optimization.
  Different colors indicate distinct seeds for model training.  }
  \label{fig:optimizer_seeds}
\end{figure}

We trained the natural and robust MNIST models with five different seeds and extracted adversarial subspaces for each model to examine the influence of training seeds on our results.
Figure \ref{fig:optimizer_seeds} shows that there is slight variation in number of adversarial subspace directions found by the optimizer.
However, the overall trend of bar heights is similar across seeds and there are no significant outliers. 

\begin{figure}[ht]
  \centering
  \includegraphics{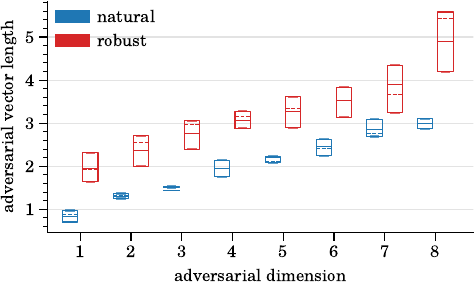}
  \caption{\textbf{Distance consistency on MNIST}.
  Box plots show the length of the adversarial vector as one increases dimensionality.
  The data is computed for a single image, where the variation comes from five different seeds used to initialize the networks before training.
  Note that the variation is much smaller than the variation across images for a single seed, as can be seen in Figure \ref{fig:mnist_perturbation_distances}.
  }
  \label{fig:mnist_seed_distances}
\end{figure}

To further validate our results, we repeated the experiment for Figure \ref{fig:mnist_perturbation_distances} (measuring the length of the adversarial vector as one increases dimensionality on the MNIST trained networks) across five different seeds per training type.
The results are shown in Figure \ref{fig:mnist_seed_distances}.
The variation across seeds for a single image is smaller than the variation across images for a single seed, which can be seen by comparing Figure \ref{fig:mnist_seed_distances} to Figure \ref{fig:mnist_perturbation_distances}.
In general, we also found that the results and trends were consistent between MNIST and CIFAR-10 networks (with differences noted in Appendix \ref{app:mnist}). 
Thus, we considered one seed per CIFAR-10 model as sufficiently representative. 

We also investigated the influence of random number generation relevant for sampling within \textit{adversarial subspaces}.
As an empirical test, we varied the number of samples and measured the change in mean distance to the decision boundary as one increases the number of samples.
This was repeated for 100 different 50-dimensional subspaces (i.e. subspaces for 100 different test images) with the naturally trained CIFAR network.
Figure \ref{fig:sample_convergence} shows that the absolute change decreases steadily to an average l2 distance of 0.004.
Thus, the mean variation from 100 samples is much smaller than the general distance scales we are measuring throughout the paper.
We also measured the variation in distances across random seeds for the sampling process.
To do this, we used 10 different seeds to generate 100 samples in a 50-dimensional subspace corresponding to one test image.
For each seed, we compute a mean distance, averaged across 100 samples, giving us 10 values.
The mean distance across seeds was 0.0933, and the standard deviation was 0.0007.
The standard deviation is less than 1\% of the mean and thus sufficiently small to consider an estimate with 100 samples as representative.

\section{Comparisons to related work: Orthogonality and clipping}\label{app:orthogonal_clipping}
One obstacle in finding proper adversarial subspaces spanned by \textit{orthogonal} vectors is assuring that perturbations do not cause pixel values to be out of the valid input boundaries $[0,1]$. 
We noticed that many studies solve this problem by element wise pixel clipping.
Here, we show that pixel clipping is problematic when talking about the dimensionality of subspaces.
Element wise pixel clipping is a non-linear operation performed on a vector and can therefore change its direction in the input space and its relative angle to other, fixed vectors.

To show that clipping has a non-negligible effect, we investigated relative angles of basis vectors that span (sub-) spaces for 100 MNIST test images relevant for two studies. 

\begin{figure}[ht]
  \vspace{-5pt}
  \centering
  \includegraphics[width=\textwidth]{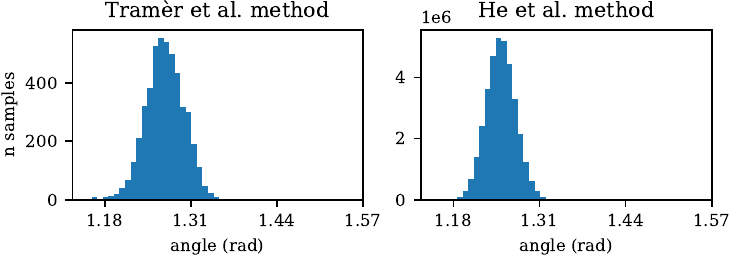} 
  \caption{\textbf{Clipping causes non-orthogonality}. The histograms show angles between vectors that define ``subspaces'' in the GAAS method proposed by \cite{tramer2017space} (left) and  distance to decision boundary estimations by \cite{he2018decision} (right).}
  \label{fig:angles}
\end{figure}

First, in their work, \cite{tramer2017space} propose the GAAS method.
GAAS estimates the dimensionality of adversarial subspaces by defining an orthogonal set of vectors aligned with an attack vector to an input image.
However, before passing the orthogonal basis vectors to the model, a clipping operation is performed.
Figure \ref{fig:angles} (left) shows that this clipping operation causes the input vectors to be non-orthogonal -- orthogonality would constitute an angle of 90° or $\frac{\pi}{2}$.
In fact, for the 100 sample images we tested, there is not a single vector pair that is completely orthogonal.
The author's conclusions regarding the dimensionality of adversarial subspaces and transferable subspaces are therefore grounded on ill-defined subspaces, spanned by non-orthogonal vectors.

In another study, \cite{he2018decision} estimate the distance to the decision boundary for the complete input space of MNIST.
For that, a  basis of 784 vectors, spanning the whole input space, is randomly generated.
Again, the inputs are clipped before they are passed to the model leading to the same problem as described earlier.
And as before, there are no orthogonal vector pairs, and angels are accumulated at much smaller values than $\frac{\pi}{2}$as can be seen in Figure \ref{fig:angles} (right).
We therefore note that it is not guaranteed that the input space is evenly covered and that edges of the 784-dimensional hypercube (in the case of MNIST) are arguably overrepresented.

\section{Comparisons to related work: Decision boundary curvature}\label{app:moosavi_comparison}
\begin{table}[ht]
    \centering
    \begin{tabular}{c c c c c c}
        \toprule
         Reference & Derivation & Generating matrix & $P$ & Dimension & \\\midrule
         \citealp{poole2016exponential}      
         & None & $\frac{P\hess_{F_{ij}}P}{\norm{\nabla F_{ij}}_{2}}$  
         & $\id - \frac{\nabla F_{ij}\nabla F_{ij}^\trsp}{\norm{\nabla F_{ij}}^2}$ 
         & ${n\times n}$ &  \\
         \citealp{moosavi2018robustness}     
         & None & $P\hess_{F_{ij}}P$                                 
         & \text{Not given}                
         & ${n\times n}$ &  \\
         \citealp{moosavi2019robustness}     
         & None & \text{Crossenetropy Hessian}                            
         & - & ${n\times n}$ & \\
         \citealp[p.~90]{moosavi2019geometry} 
         & None & $\frac{P\hess_{F_{ij}}P}{\norm{\nabla F_{ij}}_2}$  
         & $\id - \nabla F_{ij}\nabla F_{ij}^\trsp$               
         & ${n\times n}$ &  \\\midrule                            
         This work                           
         & Appendix \ref{app:curvature_long}                 
         & \eqref{eq:shape_operator} using \eqref{eq:hess_verbose} 
         & -  & ${n-1\times n-1}$ &  \\\bottomrule
    \end{tabular}
    \caption{
        Decision boundary curvature formulas present in the literature for a decision function $F_{ij}:\R^n\to \R$. 
        The \textit{generating matrix} column refers to the matrix used to yield eigenvalues for the principle curvatures, and is roughly comparable to the shape operator.
    }
    \label{tab:shape_operator_comparisons}
\end{table}

This section provides a detailed comparison of our analysis against previously published works on the principal curvature analysis applied to decision boundaries.
We include table \ref{tab:shape_operator_comparisons} for a quick summary.

The use of differential geometry to understand the non-linear response properties of model neurons has a long history in computational neuroscience, with early contributions made by \citet{zetzsche1990image}.
More recently, several works have specifically used Reimannian geometry to measure the principal curvature of response manifolds \citep{fawzi2016robustness, poole2016exponential, moosavi2018robustness, moosavi2019robustness, golden2019selective}.
As we discuss in Appendix \ref{app:curvature_long}, certain smoothness assumptions are required for such an analysis.
Our present study, as well \citet{poole2016exponential} and \citet{golden2019selective}, avoid this problem by using twice-differentiable activation functions.
In the case of \citet{golden2019selective}, this was done without retraining the network, while \citet{poole2016exponential} used nets with random weights.
We instead trained the twice-differentiable networks from scratch.
Moosavi-Dezfooli et al. used a finite-difference approximation to avoid the problem \citep[described in][eq 1 and surrounding discussion]{moosavi2019robustness}.
Another high-level difference, discussed in detail in Appendix \ref{app:curvature_long}, is the choice of measuring the response graph curvature (which is $n$ dimensional for $n$ dimensional inputs) or the level-set curvature (which is $n-1$ dimensional for $n$ dimensional inputs) of a given function.
This difference was discussed in \citep[][, at the end of Section 4 and in Figure 4]{golden2019selective}, although their technical derivations are limited only to the graph formulation.
The other studies did not explicitly discuss this difference.
Importantly, though, the decision boundary is the level-set of the logit difference function, as defined in Equation \ref{eq:boundary_def}.

Several works from Moosavi-Dezfooli and colleagues explore decision boundary curvature and its relationship to adversarial robustness \citep{fawzi2016robustness, moosavi2017analysis, moosavi2018robustness, moosavi2019robustness, moosavi2019geometry}.
Within these works they provide an inspired perspective on explaining adversarial robustness in terms of the curvature of decision boundaries.
Almost all of the work is with respect to \emph{universal} adversarial perturbations, which share a common subspace \citep{moosavi2017universal}, although \citet{fawzi2016robustness} alternatively look at random subspaces.
Their universal adversarial perturbation subspaces differ from our adversarial subspaces in that theirs are defined by orthogonal directions that transfer well across inputs instead of those that point to the nearest decision boundary for a given input.
Despite these differences, the stated quantity of interest, that is, the principle curvatures at a point on the decision boundary is the same in their work as well as ours.

However, each of these papers have a slightly different analytic formula for estimating the curvature of the boundary within a given subspace \citep{fawzi2016robustness, moosavi2017analysis}, or in the high-dimensional input space \citep{moosavi2018robustness, moosavi2019robustness}.
We found the most complete description to be given in \citet{moosavi2019geometry}, which summarizes a majority of the earlier work and applies the decision boundary curvature analysis to both subspaces and the high-dimensional input space.
Therein, \citet[][Section. 7.4]{moosavi2019geometry} state the following formula, without derivation, for the shape operator (in the notation of \autoref{app:curvature_long}):
\begin{equation}\label{eq:moosavi_shape}
    s = \frac1{\norm{\nabla F_{ij}(x)}_{2}}P\hess_{F_{ij}}P,
\end{equation}
where $\hess_{F_{ij}}$ is the Hessian of the boundary function, $F_{ij}(x) \defeq f_{i}(x) - f_{j}(x)$, and $P \defeq \id - \nabla F_{ij}(x) \nabla F_{ij}(x)^\trsp$ is described as a projection operator.
This formulation is similar to that used by \citep{poole2016exponential}, with the only difference being that they use normalized gradients in the projection operator.

Our formulation differs from the above in several ways, the most important of which is that they claim to transform the Hessian by projecting it onto the tangent space (as written above), while we apply the implicit function theorem to parameterize the decision boundary as the graph of a function (see Appendix Section \ref{sec:level_set_method}, Equation \ref{eq:implicit_function_theorem}).
For computing the curvature of the decision boundary, exclusively projecting the Hessian is not going to result in the correct calculations, since for the decision boundary (as defined in Section \ref{sec:curvature_short}, Equation \ref{eq:boundary_def}) you need a different normal than for the graph of the response function.
More specifically, the normal used for computing the shape operator of the graph surface is different from that for computing the shape operator of the decision boundary.
The latter has to be exclusively in the pixel space, while the former always has a component in the direction of the function response.
Projecting the Hessian can restrict the computed curvatures in a certain subspace of directions, but it cannot change the computed curvatures in said subspace.
Another important difference is their projection operator is computed directly from the gradient, and thus has a dimensionality equal to the input dimensionality.
To calculate the curvature of a decision boundary, one needs a level-set formulation, which should have one fewer dimensions than the input space.

The consequence of this difference can be made clear when looking at the curvature of an $n$-dimensional sphere, as defined by the quadratic function:
\begin{equation}
    F(x) = x^{T} H x,
\end{equation}
where $x$ is the input point vector and $H$ is the sphere's Hessian (i.e. an identity matrix in $\R^{n\times n}$).
Given this function, the principal curvatures of the level-set at some radius, $r$, are all known to be (up to a sign change) $\frac{1}{r}$.
However, we show in table \ref{tab:sphere_curvature_comparisons} that the formulations provided by \citet{moosavi2019geometry} and \citet{poole2016exponential} produced curvatures that are often several orders of magnitude off from the correct values the case of high-dimensional inputs.
It is possible that these previous works compute the correct curvature for their specific situation studied, although this would be difficult to verify without a full derivation of the curvature quantities and our sphere test is closely matched to the decision boundary level-set curvature that we measure.
Due to the dimensionality issue explained above, we computed the average error across all principal curvatures, regardless of the number of curvatures reported, as they should all equal $\frac{1}{r}$.

\begin{table}
   \centering
   \vspace{-10pt}
   \begin{tabular}{c c c c}
   \\
   \toprule
   dimensionality-radius & Moosavi et al. & Poole et al. & Ours \\
   \midrule
    02 - 00.1 & 7.6e-01 & 5.0e+00 & 0.0e+00 \\
    02 - 01.0 & 2.4e+01 & 5.0e-01 & 0.0e+00 \\
    02 - 10.0 & 3.1e+04 & 5.0e-02 & 1.3e-17 \\
    05 - 00.1 & 7.2e-01 & 2.0e+00 & 2.2e-15 \\
    05 - 01.0 & 7.2e+01 & 2.0e-01 & 2.2e-16 \\
    05 - 10.0 & 7.9e+04 & 2.0e-02 & 3.8e-17 \\
    10 - 00.1 & 6.4e-01 & 1.0e+00 & 2.3e-15 \\
    10 - 01.0 & 1.5e+02 & 1.0e-01 & 2.2e-16 \\
    10 - 10.0 & 1.5e+05 & 1.0e-02 & 4.4e-17 \\
   \bottomrule
   \end{tabular}
   \caption{Calculations from alternative attempts find incorrect principal curvature for a level-set on a sphere of varying dimensionality and radius. The quantity reported is the mean of the absolute difference between each computed principal curvature and the inverse radius.}
   \label{tab:sphere_curvature_comparisons}
\end{table}

\begin{table}
   \centering
   \vspace{-10pt}
   \begin{tabular}{c c}
   \\
   \toprule
   Analysis type & Absolute error \\
   \midrule
   Moosavi et al. & 6.8e+00 \\
   Poole et al. & 1.4e-03 \\
   Golden et al. & 4.1e-10 \\
   Ours & 4.1e-10 \\
   \bottomrule
   \end{tabular}
   \caption{Calculations from alternative attempts find incorrect Gaussian curvature (i.e. the product of the principal curvatures) for a Hyperboloid graph.}
   \label{tab:graph_curvature_comparisons}
\end{table}

We additionally tested a graph formulation (defined in Section \ref{sec:graph_manifolds}), using a one-sheeted hyperboloid.
This formulation lives in the full space, so there is a principal curvature per input dimension.
However, we show similar errors in in Table \ref{tab:graph_curvature_comparisons}.
Note, that the formulation from \citep{golden2019selective} matched our own, although they did not provide an equation for mapping this to the level-set curvature.
Our own implementations were correct with high precision on these and other experiments with known functions, all of which can be replicated from our provided code.
Finally, while the subspace projection methods proposed by \citep{fawzi2016robustness, moosavi2017analysis} potentially provide metrics related to curvature in the subspaces they defined, we chose to instead derive our pullback method described in Equation \ref{eq:subspace_pullback} from primary sources, as explained in Appendix \ref{app:curvature_long}.


\end{document}